\documentclass[acmsmall]{acmart}

\usepackage{hyperref} 
\usepackage{graphicx}
\usepackage{booktabs}
\usepackage{amsmath}
\usepackage{amsthm}
\usepackage{bbm}
\usepackage{amsfonts}
\usepackage{epsfig}
\usepackage{graphicx}
\usepackage{subfigure}
\usepackage{url}
\usepackage{diagbox}
\usepackage{balance}
\usepackage[noend]{algorithmic}
\usepackage{algorithm}
\usepackage{multirow}
\usepackage{xspace}

\usepackage{pifont}
\newcommand{\cmark}{\ding{51}}%
\newcommand{\xmark}{\ding{55}}%

\newtheorem{problem}{Problem}

\newcommand{\spara}[1]{\smallskip\noindent{\bf #1}}

\newcommand{\squishlist}{
 \begin{list}{$\bullet$}
  {  \setlength{\itemsep}{0pt}
     \setlength{\parsep}{3pt}
     \setlength{\topsep}{3pt}
     \setlength{\partopsep}{0pt}
     \setlength{\leftmargin}{2em}
     \setlength{\labelwidth}{1.5em}
     \setlength{\labelsep}{0.5em}
} }
\newcommand{\squishlisttight}{
 \begin{list}{$\bullet$}
  { \setlength{\itemsep}{0pt}
    \setlength{\parsep}{0pt}
    \setlength{\topsep}{0pt}
    \setlength{\partopsep}{0pt}
    \setlength{\leftmargin}{2em}
    \setlength{\labelwidth}{1.5em}
    \setlength{\labelsep}{0.5em}
} }

\newcommand{\squishdesc}{
 \begin{list}{}
  {  \setlength{\itemsep}{0pt}
     \setlength{\parsep}{3pt}
     \setlength{\topsep}{3pt}
     \setlength{\partopsep}{0pt}
     \setlength{\leftmargin}{1em}
     \setlength{\labelwidth}{1.5em}
     \setlength{\labelsep}{0.5em}
} }

\newcommand{\squishend}{
  \end{list}
}

\renewcommand{\And}{\mbox{\bf and}\ }

\newcommand{\False}{\mbox{\em false}}
\newcommand{\True}{\mbox{\em true}}

\newcommand{\eat}[1]{}

\newcommand{\ie}{i.e.,\xspace}
\newcommand{\eg}{e.g.,\xspace}
\newcommand{\wrt}{w.r.t.\xspace}

\newcommand{\NP}{\ensuremath{\mathbf{NP}}\xspace}

\newcommand{\kw}[1]{{\ensuremath {\mathsf{#1}}}\xspace}

\newcommand{\stitle}[1]{\vspace{1.5ex}\noindent{\bf #1}}
\newcommand{\etitle}[1]{\vspace{0.8ex}\noindent{\em #1}}
\newcommand{\eetitle}[1]{\vspace{0.8ex}\noindent{\em\underline{#1}}}

\newcommand{\sstab}{\rule{0pt}{8pt}\\[-2.4ex]}
\newcounter{ccc}

\DeclareMathOperator*{\argmax}{arg\,max}

\usepackage[show]{boxnotes}
\usepackage[normalem]{ulem}
\newcommand\redout{\bgroup\markoverwith
{\textcolor{red}{\rule[.5ex]{2pt}{2pt}}}\ULon}

\renewcommand{\P}{{\mathcal P}}
\newcommand{\G}{{\mathcal G}}
\newcommand{\V}{{\mathcal V}}
\newcommand{\X}{{\mathcal X}}

\newcommand{\gvl}{$\G^l_{\mathcal V}$\xspace}

\newcommand{\M}{{\mathcal M}}
\newcommand{\C}{{\mathcal C}}
\newcommand{\gnn}{\kw{GNN}}
\newcommand{\gnns}{\kw{GNNs}}

\newcommand{\gcn}{\kw{GCN}}

\newcommand{\evg}{\kw{EVG}}
\newcommand{\gnninf}{\kw{EVerify}}
\newcommand{\pmatch}{\kw{PMatch}}

\newcommand{\vpexp}{\kw{VpExtend}}

\newcommand{\psum}{\kw{Psum}}
\newcommand{\pgen}{\kw{PGen}}

\newcommand{\PTIME}{\kw{PTIME}}

\newcommand{\gvex}{\kw{GVEX}}

\newcommand{\gvapprox}{\kw{ApproxGVEX}}
\newcommand{\gvstream}{\kw{StreamGVEX}}

\renewenvironment{proof}{
        \vspace{1ex}
        {\noindent\bf Proof:}}{\vspace{1ex}}
\newenvironment{proofS}{
        \vspace{1ex}
        {\noindent\bf Proof sketch:\ }}{\vspace{1ex}}

\renewcommand{\st}[1]{}
\usepackage[tikz]{bclogo}

\AtBeginDocument{%
  }

\setcopyright{acmlicensed}

\acmJournal{PACMMOD}
\acmYear{2024} 
\acmVolume{2} 
\acmNumber{1 (SIGMOD)}
\acmArticle{40}
\acmMonth{2} 
\acmDOI{10.1145/3639295}

\begin{document}

\title{View-based Explanations for Graph Neural Networks}

\author{Tingyang Chen}
\authornote{Both authors contributed equally to this research.}
\email{chenty@zju.edu.cn}
\orcid{0009-0008-5635-9326}
\affiliation{%
  \institution{Zhejiang University}
  \city{Ningbo}
  \country{China}
}

\author{Dazhuo Qiu}
\authornotemark[1]
\orcid{0000-0002-1044-5252}
\email{dazhuoq@cs.aau.dk}
\affiliation{%
  \institution{Aalborg University}
  \city{Aalborg}
  \country{Denmark}}

\author{Yinghui Wu}
\email{yxw1650@case.edu}
\orcid{0000-0003-3991-5155}
\affiliation{%
  \institution{Case Western Reserve University}
  \city{Cleveland}
  \state{Ohio}
  \country{USA}
}

\author{Arijit Khan}
\email{arijitk@cs.aau.dk}
\orcid{0000-0002-7312-6312}
\affiliation{%
 \institution{Aalborg University}
 \city{Aalborg}
 \country{Denmark}}

\author{Xiangyu Ke}
\orcid{0000-0001-8082-7398}
\email{xiangyu.ke@zju.edu.cn}
\affiliation{%
  \institution{Zhejiang University}
  \city{Hangzhou \& Ningbo}
  \country{China}
}

\author{Yunjun Gao}
\email{gaoyj@zju.edu.cn}
\orcid{0000-0003-3816-8450}
\affiliation{%
  \institution{Zhejiang University}
  \city{Hangzhou}
  \country{China}
}

\renewcommand{\shortauthors}{Tingyang Chen, et al.}

\begin{abstract}
Generating explanations for graph neural networks (\gnns) has been studied to understand their behaviors in analytical tasks such as graph classification. Existing approaches aim to understand the overall results of \gnns rather than providing explanations for specific class labels of interest, and may return explanation structures that are hard to access, nor directly queryable. 
We propose \gvex, a novel paradigm that generates  \underline{\textbf{G}}raph \underline{\textbf{V}}iews for GNN \underline{\textbf{EX}}planation.  
{\bf (1)} We design a two-tier explanation 
structure called {\em explanation views}. An explanation view consists of a set of graph patterns and a set of induced explanation subgraphs. Given a database $\G$ of multiple graphs and a specific class label $l$ assigned by a GNN-based classifier $\M$, it concisely describes the fraction of $\G$ that best explains why $l$ is assigned by $\M$. 
{\bf (2)} We propose quality measures and formulate an optimization problem to compute optimal explanation views for \gnn explanation. We show that the problem is  $\Sigma^2_P$-hard. 
{\bf (3)} We present two algorithms. 
The first one follows an {\em explain-and-summarize} strategy that first generates high-quality explanation subgraphs which best explain \gnns in terms of feature influence maximization, and then performs a summarization step to generate patterns. We show that this strategy provides an approximation ratio of $\frac{1}{2}$. 
Our second algorithm performs a single-pass to an input node stream in batches to incrementally maintain explanation views, having an anytime quality guarantee of $\frac{1}{4}$-approximation. Using real-world benchmark data, we experimentally demonstrate the effectiveness, efficiency, and scalability of \gvex. Through case studies, we showcase the practical applications of \gvex.

\end{abstract}



\received{July 2023}
\received[revised]{October 2023}
\received[accepted]{November 2023}

\maketitle

\vspace{-0.5mm}
\section{Introduction}
\label{sec:intro}
Graph classification is 
essential for a number of real-world tasks such as drug discovery, text classification, and recommender system~\cite{hu2020open, yao2019graph, wu2022graph}. The rising graph neural networks  (\gnns) have exhibited great potential in graph classification across many real domains, e.g., social networks, chemistry, and biology~\cite{zitnik2018modeling, you2018graph, cho2011friendship,wei2023neural}. 
Given a database $\G$ as a set of graphs,  and a set of class labels $\L$, \gnn-based 
graph classification aims to learn a \gnn 
as a classifier $\M$, such that each graph 
$G\in \G$ is assigned a correct label $\M(G)\in \L$. 

Nevertheless, it remains a desirable yet nontrivial task to explain the results of high-quality \gnn-classifiers for domain experts~\cite{yuan2022explainability}. Given $\M$ and $\G$, one wants to discover a critical fraction of $\G$ that is responsible for the occurrence of specific class labels of interest, assigned by \gnn $\M$ over $\G$. 
In particular, such explanation structures should 
{\bf (1)} capture both important features and topological structural information in $\G$; 
{\bf (2)} be {\em queryable}, hence are easy for human experts to access and inspect with domain knowledge.  

Existing \gnn explanation techniques 
\cite{ying2019gnnexplainer,yuan2021explainability,huang2023global,zhang2022gstarx,yuan2020xgnn} primarily characterize 
explanations as important 
input features (typically in the form of numerical encodings) 
directly from \gnn layers, and remain limited to retrieving structural information as needed~\cite{yuan2022explainability}. 
These feature encodings alone 
cannot easily express ``queryable'' substructures such as subgraphs and graph patterns~\cite{yuan2021explainability}. Indeed, 
graph patterns are often more intuitive to relate useful functionalities and better bridge human knowledge with \gnn results.
Moreover, the generated explanations typically aim to clarify all assigned labels rather than specifically addressing user-specified class labels of interest. Consider the following real-world example. 
\begin{figure}[tb!]
\vspace{-1mm}
\centering
\centerline{\includegraphics[scale=0.28]{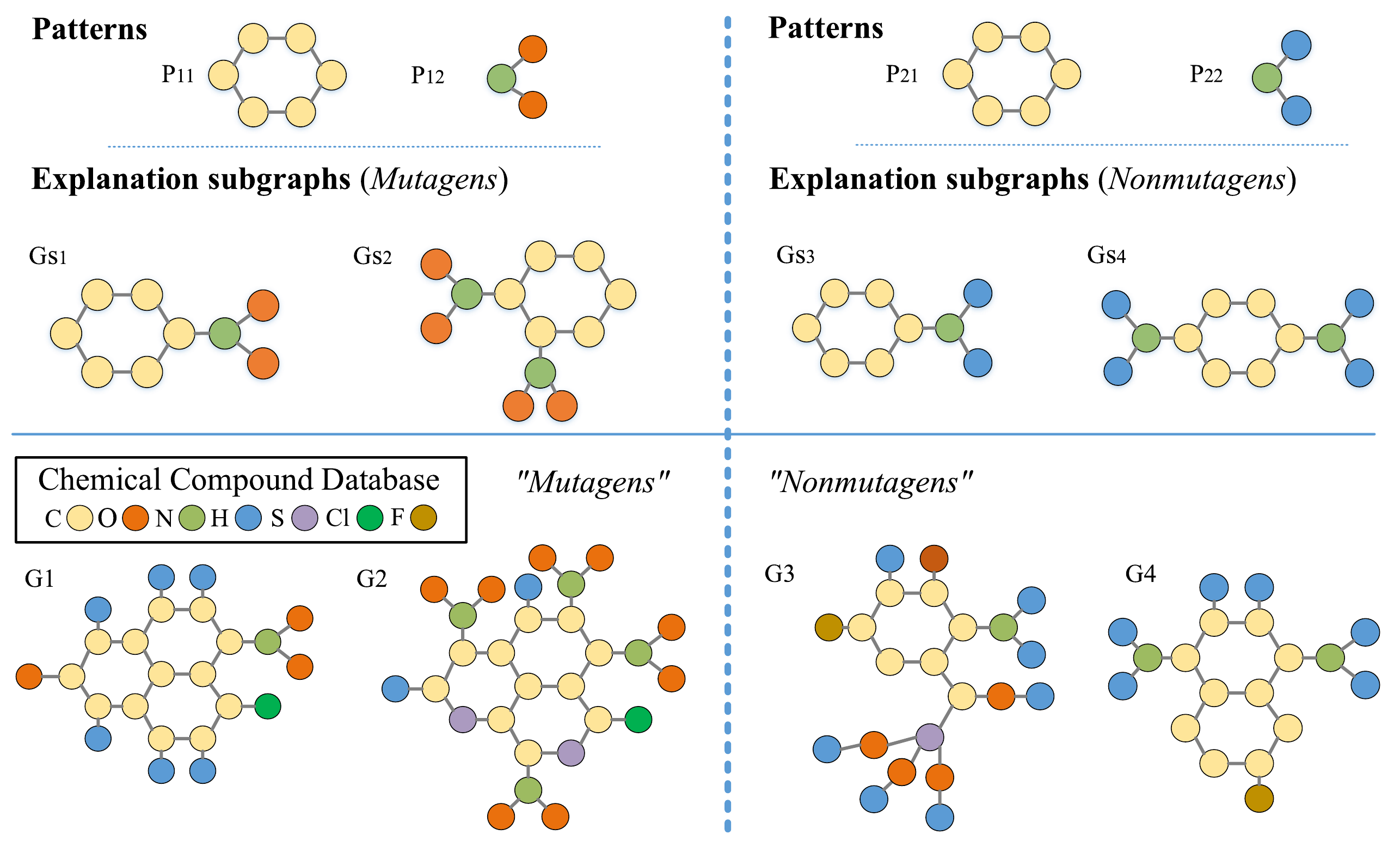}}
\vspace{-3mm}
\caption{\gnn-based drug classification, with patterns and induced subgraphs 
that help understand the results.}
\label{fig-motivation}
\vspace{-2mm}
\end{figure}
\begin{table*}[tb!]
    \centering
     \caption{\small Comparison of our \gvex technique with state-of-the-art \gnn explanation methods. Here ``Learning'' denotes whether (node/edge mask) learning is required, ``Task'' means what downstream tasks each method can be applied to (GC/NC: graph/ node classification), ``Target'' indicates the output format of explanations (E/NF: Edge/Node Features), ``Model-agnostic''(MA) means if the method treats \gnns as a black-box during the explanation stage (i.e., the internals of the \gnn models are not required), ``Label-specific"(LS) means if the explanations can be generated for a specific class label; ``Size-bound''(SB) means if the size of explanation is bounded; ``Coverage'' means if the coverage property is involved (\S \ref{sec-view}), ``Config'' means if users can configure the method to generate explanations for 
     designated class labels (\S \ref{sec:preliminaries}); ``Queryable'' means if the explanations are 
     directly queryable.} 
    \label{tab:comprehensive-analysis}
    \vspace{-2mm}
    \scriptsize
    \begin{tabular}{c c c c c c c c c c}
            \textbf{Methods} & \textsc{Learning} & \textsc{Task} & \textsc{Target} & \textsc{MA} & \textsc{LS} & \textsc{SB} & \textsc{Coverage} & \textsc{Config} & \textsc{Queryable} \\ 
            \midrule
		  {\bf SubgraphX}~\cite{yuan2021explainability} & \xmark & GC/NC & Subgraph & \cmark & \xmark & \xmark & \xmark & \xmark & \xmark\\ 
            {\bf GNNExplainer}~\cite{ying2019gnnexplainer} & \cmark & GC/NC & E/NF & \cmark & \xmark & \xmark & \xmark & \xmark & \xmark\\
            {\bf PGExplainer}~\cite{luo2020parameterized} & \cmark & GC/NC & E & \xmark & \xmark & \xmark & \xmark & \xmark & \xmark\\
            {\bf GStarX}~\cite{zhang2022gstarx} & \xmark & GC & Subgraph & \cmark & \xmark & \xmark & \xmark & \xmark & \xmark\\
            {\bf GCFExplainer}~\cite{huang2023global} & \xmark  & GC & Subgraph & \cmark & \cmark & \xmark & \cmark & \xmark & \xmark\\
            \midrule
            {\bf \gvex (Ours)} & \xmark  & GC/NC & \begin{tabular}{@{}c@{}}Graph Views \\ (Pattern+Subgraph)\end{tabular} & \cmark & \cmark & \cmark & \cmark & \cmark & \cmark \\
    \end{tabular}
    \vspace{-3mm}
\end{table*}
\begin{example}
\label{exa-motivation}
In drug discovery, {\em mutagenicity} 
refers to the ability of a chemical compound to cause mutation. It is an adverse property of a molecule that hampers its potential to become a marketable drug 
and has been of great interest in the field.  An emerging 
application of \gnns 
is to classify chemical compound 
as graphs in terms of mutagenicity to 
support effective drug discovery~\cite{xiong2021graph, jiang2020drug}. 

Consider a real-world molecular dataset represented as graphs in Figure \ref{fig-motivation}. 
A \gnn classifies four graphs $\{G_1, G_2, G_3, G_4\}$ into two groups with class labels ``Mutagens'' and 
``Nonmutagens'', respectively, based on whether they have mutagenicity property. A medical analyst 
seeks to understand ``why'' in particular the first two  
chemical compounds $\{G_1, G_2\}$ are recognized as mutagens, ``what'' are critical molecular substructures 
that may lead to such classification results, and further wants to search for  
and compare the difference between 
these compounds that contribute to their mutagenicity using domain knowledge. The large number of 
chemical graphs 
makes a direct inspection of \gnn results challenging. For example, it is difficult to discern whether the binding of multiple carbon rings or the presence of hydrogen atoms on the carbon rings plays a decisive role in \gnn-based classification to decide mutagenicity. 

Based on domain knowledge, {\em toxicophores} are substructures of 
chemical compounds that indicate an increased potential for mutagenicity. For example, the aromatic nitro group is a well-known toxicophore for mutagenicity ~\cite{kazius2005derivation}. 
Such structures can be readily encoded as 
``graph views'' with a two-tier structure, 
where toxicophores are expressed as graph patterns that summarize  common {sub}structures of a set of ``explanation'' subgraphs, as illustrated in 
Figure \ref{fig-motivation}. 
The upper left corner of the figure shows two graph patterns \{$P_{11}$, $P_{12}$\} and corresponding subgraphs that explain ``why'' the compounds $G_1$ and $G_2$ have mutagenicity. Indeed, we find that  
(1) if these subgraphs are removed 
from $G_1$ and $G_2$, the remaining 
part can no longer be recognized by the same \gnn as mutagens; and 
(2) two of the patterns $P_{11}$ 
and $P_{12}$ are real toxicophores as verified by domain experts. Similarly, the middle right corner depicts subgraphs with common structures summarized by 
graph patterns $P_{21}$ to $P_{22}$, which are responsible for nonmutagens. Surprisingly, $P_{21}$ and $P_{22}$ are also toxicophores as per domain knowledge. 
Therefore, such patterns can be suggested to the analysts for further inspection, or be conveniently issued as graph queries for downstream analysis, e.g., ``{\em which toxicophores occur in mutagens?'' ``Which nonmutagens contain the toxicophore pattern $P_{22}$''?}. 

In addition, the analyst wants to 
understand representative 
substructures that are 
discriminative enough to 
distinguish mutagens and nonmutagens. This can be captured by the specific toxicophore 
$P_{12}$ that covers the nodes in all two chemical compounds $\{G_1, G_2\}$ with label ``mutagens'', but does not occur in 
nonmutagens $\{G_3, G_4\}$. 
These graph patterns, along with their matched subgraphs, provide concise and queryable structures to humans, enabling a more intuitive understanding of the \gnn-based mutagenicity analysis.
\end{example}

The above example illustrates 
the need to generate 
queryable explanation 
structures which can effectively describe 
the fraction of graph data
that are responsible for the 
occurrence of user-specified class labels 
in \gnn-based classification. 
A desirable paradigm would {\bf (1)} be 
{\em model-agnostic}, {\ie does not require internals of the \gnns to generate explanations}; 
{\bf (2)} be {\em specific} in distinguishing 
the explanations for different class labels;  {\bf (3)} be {\em representative} to cover   
important substructure 
of the graphs that are 
assigned with the labels of interests, 
without over- or under-representing them (formally stated in \S \ref{sec-view}); 
{\bf (4)} be {\em configurable} to enable users with the flexibility to freely select a designated number of nodes from different classes, to obtain comprehensive and detailed explanations tailored to their classes of interest (\S \ref{sec:preliminaries}); and {\bf (5)} be 
{\em queryable} to provide direct access for human experts with (domain-aware) queries. 
None of the existing \gnn explanation 
methods can address these desirable properties (Table~\ref{tab:comprehensive-analysis}). 

\vspace{.5ex}
Graph views and view selection have been studied as an effective way to access graph data~\cite{fan2014answering,zhang2021automatic,mami2012survey}. 
Given a graph $G$, a {\em graph view} contains a
graph pattern $P$ and a materialized
subgraph $P(G)$ that matches $P$ via graph pattern matching. 
We advocate that graph views, as two-tier explanation structures, fit naturally 
to explain \gnn-based classification. Indeed, the subgraphs possess the ability to describe the essential structure of original graphs in a manner that is both model-agnostic and configurable. The ability to configure our 
\gvex algorithms by ensuring a desirable amount of nodes from each class label to
be covered, enables domain-expert users to extract more relevant
information for their specific inquiries, as presented in our case study (\S\ref{sec:casestudy}). Consequently, these subgraphs inherently exhibit discriminative and informative properties for distinct classes. To enhance user inspection, we introduce patterns as a queryable structure through pattern mining, it is a summary of subgraphs and it helps domain experts inspect large-size explanations based on their higher-tier patterns, thereby facilitating easy access and analysis. We are 
interested in the following two questions: 
{\bf (1)} {\em How to characterize \gnn 
explanation with graph views?} and 
{\bf (2)} {\em how can we generate graph views to extract explanations for \gnn in a concise and configurable manner?}
The answers to these questions 
not only provide new perspectives 
towards explainability, but also enable finer-grained, 
class label-specific analysis of \gnns. 

\vspace{1mm}
\stitle{Contributions.} 
We summarize our main contributions 
as follows. 

\spara{(1)} We introduce {\em explanation views}, a novel class of explanation structure for \gnn-based 
graph classification. 
An explanation view is a two-tier  
structure that consists of graph patterns 
and a set of explanation 
subgraphs induced from 
graphs via graph pattern matching, such that 
(a) the subgraphs are responsible 
for the occurrence of specific class label $l$ of user's interest,  
and 
(b) the patterns 
summarize the details 
of explanation subgraphs as common 
substructures for efficient 
search and comparison of 
these subgraphs (\S\ref{sec:preliminaries}). 

\spara{(2)} 
For explanation views, we introduce
a set of quality measures in terms of 
explainability 
and coverage
properties (\S\ref{sec-view}). 
We formulate the problem 
to compute the optimal explanation 
views for \gnn-based graph classification. 
The problem is in general
$\Sigma^2_P$-hard, and even
remains $NP$-hard for 
a special case when 
$\G$ has no edge.  

\spara{(3)} We present \gvex, an algorithmic 
solution 
for generating 
graph views to explain \gnns. 
(a) We first introduce an approximation scheme 
(\S\ref{sec-approx}) that  
follows an ``explain-and-summarize'' strategy. The method first computes a set of 
node-induced subgraphs with guaranteed explainability, 
by identifying important nodes 
with the maximum diversified feature influence.  
We then summarize these subgraphs 
into graph patterns 
that ensures to cover all such nodes, and 
meanwhile, introduce small edge coverage error. 
This guarantees an overall $\frac{1}{2}$-approximation
for 
the view generation problem. 
(b) We further develop a streaming 
algorithm (\S\ref{sec-stream}) that 
avoids generation of all 
explanation 
subgraphs. 
The algorithm 
processes a batched stream of 
nodes 
and incrementally 
maintains explanation views  
under the coverage constraint, with 
an approximation ratio $\frac{1}{4}$ 
relative to the processed fraction of 
graphs. 

\spara{(4)} Using real-world graphs and 
representative \gnns, we 
experimentally 
verify that our view generation 
algorithms can effectively 
retrieve and summarize substructures 
to explain 
\gnn-based classification (\S\ref{sec:exp}). 
We also showcase that our algorithms can
support \gnn-based drug design and 
social analysis well.

\stitle{Related Work}.
We summarize the related work as follows. 

\etitle{Graph Neural Networks}. 
Graph neural networks (\gnns) are deep learning models designed to tackle graph-related tasks in an end-to-end manner~\cite{wu2020comprehensive}. \st{Graphs are widely used to represent data in various real-world domains, and GNNs have exhibited promising performance on such data.} 
While \gnns 
have several variants (\eg graph convolutional networks ({\sf GCNs})~\cite{kipf2016semi}, graph attention networks ({\sf GATs})~\cite{velivckovic2017graph}, graph isomorphism networks ({\sf GINs})~\cite{xu2018how}, {\sf APPNP}~\cite{gasteiger2018predict}, and {\sf GraphSAGE}~\cite{hamilton2017inductive}), they 
share a similar feature learning paradigm: For each node, \gnns update the features of a node by aggregating the counterparts from its neighbors to update its own features. \gnns have demonstrated their efficacy 
on various tasks, including node and graph classification~\cite{kipf2016semi,xu2018how, zhang2018end, ying2018hierarchical}, link prediction~\cite{zhang2018link}.

\etitle{Explanation of GNNs}. 
Several approaches have been proposed to generate explanations for 
\gnns\cite{zeiler2014visualizing, schwarzenberg2019layerwise, huang2022graphlime, schlichtkrull2020interpreting, luo2020parameterized, ying2019gnnexplainer, yuan2021explainability, yuan2020xgnn, huang2023global,zhang2022gstarx}. 
Instance-level methods provide input-dependent explanations for each test graph, whereas model-level methods provide a global understanding of \gnns without considering specific input instances or class labels~\cite{yuan2022explainability}. 
For example, {\sf GNNExplainer}~\cite{ying2019gnnexplainer} learns to optimize soft masks for edges and node features to maximize the mutual information between the original and new predictions and induce important substructures from the learned masks. 
{\sf SubgraphX}~\cite{yuan2021explainability} explains {\sf GNN} models by identifying important subgraph for an input graph. It employs Shapley values to measure a subgraph's importance by considering the interactions among different graph structures. 
{\sf XGNN}~\cite{yuan2020xgnn} aims to explore high-level explanations of \gnns by generating graph patterns to maximize a specific prediction. 
{\sf GCFExplainer}~\cite{huang2023global} studies the global explainability of {\sf GNNs} through counterfactual reasoning. Specifically, it finds a small set of representative counterfactual graphs that explain all the input graphs rather than label-specific classes of users' interests.

However, these methods do not explicitly support configurable and queryable explanation structures, and are not optimized to generate explanations for user-specific labels. Meanwhile, they cannot be easily extended to support such constraints. First, achieving configurable property is computationally hard (Theorem \ref{thm-hardness}), existing solutions do not address such needs, and extending them for configurability is non-trivial due to the computational hardness. Second, the “queryable” property involves extracting commonalities within explanations to facilitate direct access to critical insights, current methods generate large explanations for label explanations and stll lack the ability to
include important patterns, hindering the efficient and effective computation of queryable structures. None of the prior methods supports all desirable properties at the same time, as illustrated in Table \ref{tab:comprehensive-analysis}. Consequently, there is a need for more effective and efficient methods for explaining GNNs that can provide interpretable and accurate explanations of their predictions.

\etitle{\st{Subset selection with group fairness.}} \st{Previous research has investigated subset selection with fairness constraints 
These investigations involve identifying a diverse subset that satisfies individual cardinality constraints for each group, given a universal set and a set of groups. To address this problem, approximation algorithms have been proposed for both max-sum and max-min diversification, as demonstrated by. The research has also explored submodular maximization under fairness constraints for data streams, presenting approximations with constant factors. However, these methods primarily focus on set coverage properties and are not directly applicable to generating explanation views for graph data with fairness constraints. To address this issue, we present both feasible approximations and streaming-style heuristics in this paper.
}

\vspace{-3ex}
\etitle{Graph Views}. Graph views have been studied as a 
useful approach to access and query large graphs~\cite{mami2012survey}. A graph view 
consists of a graph pattern and a set of subgraphs as its 
matches via graph pattern matching. Graph views are shown to be 
effective in view-based query processing~\cite{fan2014answering}, 
summarization~\cite{song2018mining}, event analysis~\cite{zhang2020distributed}, query suggestion~\cite{ma2022subgraph}, 
data cleaning~\cite{lin2019discovering}
and data pricing~\cite{chen2022gqp}. Several approaches 
have also been developed to discover 
graph views~\cite{song2018mining,lin2019discovering}. 

\etitle{Graph pattern mining}. 
Graph pattern mining techniques 
discover frequent or other important structures within graph data~\cite{yan2002gspan, ranu2009graphsig, yan2008mining, thoma2010discriminative, iyer2018asap}, which can help constructing graph views and ensure generation of higher-tier patterns from
lower-tier explanation subgraphs
in our two-tier explanation structure. However, graph mining alone is insufficient to generate GNN explanations, e.g., consistent and counterfactual lower-tier explanations (\S\ref{sec:gnn_views}).

\vspace{.5ex}
To the best of our knowledge, this is the first work that exploits graph views to support  
queryable explanation  
for \gnn-based classification. Our approach 
is a post-hoc method 
that treats \gnns as 
black-box (hence does not require details 
from \gnns, but only the output 
from its last layer), does not 
require node/edge mask training, 
and generates explanations 
as views that are queryable, 
concise, and class label-specific, 
all in a user-configurable manner (Table~\ref{tab:comprehensive-analysis}).

\section{Preliminaries}
\label{sec:preliminaries}
\vspace{-0.5ex}

\begin{table}[tb!]
\begin{small}

\caption{Table of notations} 
\label{tab:notation}
\vspace{-3mm}
\begin{tabular}{c|c}

\textbf{Symbol} & \textbf{Meaning} \\
\hline \hline
$G$ = $(V, E)$ & Graph with nodes $V$ and edges $E$ \\
\hline 
$(X, A)$ & 
 \begin{tabular}{@{}c@{}}Feature representation of $G$:  \\ 
($X$: feature matrix; $A$: adjacency matrix)\end{tabular}
\\
\hline
$\M$; $X^k$ & A GNN-based classifier; the embedding of 
node $v$ at layer $k$ of $\M$ \\
\hline
$\G$; $\V$ & A set of graphs (graph database) for classification; node group of $\G$\\
\hline 
 $G^l_s = (V_s, E_s)$; $\G^l_s$ & \begin{tabular}{@{}c@{}} An explanation subgraph $G_s^l$ induced by nodes $V_s$ \wrt class label $l$; \\ a set of explanation subgraphs $\G^l_s$ \wrt class label $l$ \end{tabular} \\
 \hline 
 $\G^l$; $\V^l$ & Label group (of graphs with label $l$); the node set of $\G^l$\\
\hline
$P$; $\P$ & A single graph pattern $P$; a set of graph patterns $\P$ \\
\hline
\gvl $= (\P^l, \G^l_s)$ & 
 \begin{tabular}{@{}c@{}} A single explanation view with \\ a pattern set
$\P^l$ and an explanation subgraph set $\G^l_s$ 
\end{tabular}  \\ 
\hline
$\C$ = ($\theta$, r, \{[$b_l,u_l$]\}) &
\begin{tabular}{@{}c@{}} 
A configuration that specifies \\
explainability ($\theta$, r) and coverage 
constraints \{[$b_l,u_l$]\} $(l\in\L)$
\end{tabular}\\
\hline
$\G_\V$ & A set of explanation views \\
\hline
\end{tabular}
\end{small}
\vspace{-2ex}
\end{table}

We start with reviewing attributed graphs, \gnns, and graph views in \S\ref{sec:gnns}. We then introduce view-based explanations in \S\ref{sec:gnn_views}. For easy reference, important notations are summarized in Table \ref{tab:notation}.

\vspace{-1ex}
\subsection{Graph Neural Networks and 
Graph Views}
\label{sec:gnns}

\vspace{-1.5ex}
\stitle{Attributed Graphs}.  
We consider a connected 
graph $G = (V,E,T,L)$, where $V$ is the set of nodes, and $E\subseteq V\times V$ a set of edges. 
Each node $v$ carries a tuple $T(v)$ 
of \st{node} attributes (or features) and their values. Each node $v\in V$ (resp. edge $e\in E$) has a {\em type} $L(v)$ (resp. L(e)). 

\vspace{-0.8ex}
\stitle{Graph Neural Networks}. 
\gnns are a family of well-established deep learning models that extend traditional neural networks to transform graphs into proper embedding representations for various downstream analysis such as graph classification. 
In a nutshell, \gnns employ a multi-layer message-passing scheme, through which the feature representation of a node in the next layer is aggregated from its neighborhood in the current layer. 
For example, the Graph Convolutional Network (\gcn) \cite{kipf2016semi}, 
a representative \gnn model, adopts a general form of the function as:
\begin{equation}
\label{eq-prop}
    X^k = \delta(\Hat{D}^{-\frac{1}{2}} \Hat{A} \Hat{D}^{-\frac{1}{2}} X^{k-1}  \Theta^{k})
\end{equation}

Here $\Hat{A} = A + I$, where $I$ represents the identity matrix and $A$ is the adjacency matrix of graph $G$. $X^k$ indicates node feature representation in the $k$-th layer, (with $X^0=X$ a matrix of input node features), where each row $X_v$ is a vector (numerical) encoding of a node tuple $T(v)$. 
The encoding can be obtained by, e.g., 
word embedding or one-hot encoding~\cite{gardner2018allennlp}. $\Hat{D}$ represents the diagonal node degree matrix of $\Hat{A}$, $\delta(\cdot)$ is the non-linear activation function, and $\Theta^{k}$ represents the learnable weight matrix for the $k$-th layer.

\vspace{-0.8ex}
\stitle{GNN-based Classification}.
The task of graph classification is to correctly assign a categorical class label for a graph. Given a database (a set of graphs) $\mathcal{G} = \{G_1, G_2, \dots, G_m\}$
and a set of class labels $\L$, 
a {\em \gnn-based classifier}  
$\M$ of $k$ layers (1) takes as input \st{their feature representations} $G_i$=$(X_i,A_i)$ $(i\in [1,m])$, learns to generate feature representations  
$X_i^k$, converts them into class labels that best fit a set of labeled 
graphs (``training examples''), and (2) assigns, for  
each unlabeled ``test'' graph $G_i\in \mathcal{G}$, a class label $l\in\L$ 
(denoted as $\M(G_i)$ = $l$). 

\vspace{.5ex}
We aim to generate 
queryable structures that 
can also clarify the class labels 
of user's interests assigned by a 
\gnn-based classifier $\M$ over $\G$. 
To this end, 
we revisit graph patterns and views as ``building block'' 
structures for view-based \gnn explanation. 

\vspace{-0.8ex}
\stitle{Graph Patterns}. 
A {\em graph pattern} 
is a connected graph $P(V_p, E_p, L_p)$, where 
$V_p$ is 
a set of pattern nodes, 
$E_p$ is a set of pattern edges, and 
$L_p$ is a function that assigns for each 
node $v_p\in V_p$ (resp. edge $e_p\in E_p$)  a {\em type} $L(v_p)$ (resp. $L(e_p)$). 

\eetitle{Graph Pattern Matching}. We use {\em node-induced subgraph isomorphism}~\cite{floderus2015induced} to characterize graph pattern matching. Given a 
graph $G$ and a pattern $P$, there is 
a matching function $h$ between 
$P$ and $G$ if for each pattern node 
$v_p\in V_P$, (1) $h(v_p)$ is a node 
in $V$ and $L(v)$ = $L_p(v_p)$, (2) $(h(v_p), h(v_p'))$ is 
an edge in $G$ if $(v_p,v_p')\in E_p$, 
and $L(h(v_p), h(v_p'))$ = $L_p(v_p, v_p')$. We say that a graph pattern 
$P$ {\em covers} a node $v$ (resp. an edge $e$) if 
there is a matching $h$ 
such that $h(v_p)$ = $v$ (resp. $h(e_p)$=$e$)  
for some $v_p\in V_p$ (resp. $e_p\in E_p$). 
Given a set of graph patterns 
$\P$ and a set of graphs $\G$, 
we say that $\P$ {\em covers} the nodes (resp. edges) in $\G$, if for every graph 
$G\in \G$ and every node $v$ (resp. edge $e$) in $G$, 
there exists a pattern $P\in \P$, 
such 
that $P$ covers $v$ (resp. $e$). 

\vspace{-0.8mm}
\stitle{Graph Views\st{: A Revisit}}. Graph views 
have been extensively studied to support 
fast graph access and query processing~\cite{mami2012survey}. 
Given a graph database $\G$, 
we represent a graph view as a ``two-tier'' structure, denoted as $\G_\V$ = $(\P, \G_s)$, 
where 
(1) $\P$ = $\{P_1,\ldots, P_n\}$ is a set of graph patterns, and 
(2) $\G_s$ is a set of connected subgraphs of the graphs from $\G$ which are 
induced by the nodes that matches the patterns 
in $\P$ via node-induced subgraph 
isomorphism (see ``Graph Pattern Matching''). 
Note that by definition, $\P$ 
covers all the nodes in $\G_s$. 

\vspace{-0.8mm}
\stitle{Remarks}. 
We remark the difference between  
a ``type'' $L(\cdot)$ and a ``class label'' $l\in \L$. The former refers to the 
real-world entity types, as seen in, 
\eg ontologies, and are enforced to be 
consistent in graph pattern matching; and the latter refers to the task-specific class labels 
assigned by \gnn-based classifiers. For simplicity, we shall refer to 
``class labels'' as ``labels'', 
``\gnn-based classifier'' 
as ``\gnn'', and ``graph patterns'' 
as ``patterns''. 

\begin{figure}[tb!]
\vspace{-1ex}
\centering
\centerline{\includegraphics[scale=0.24]{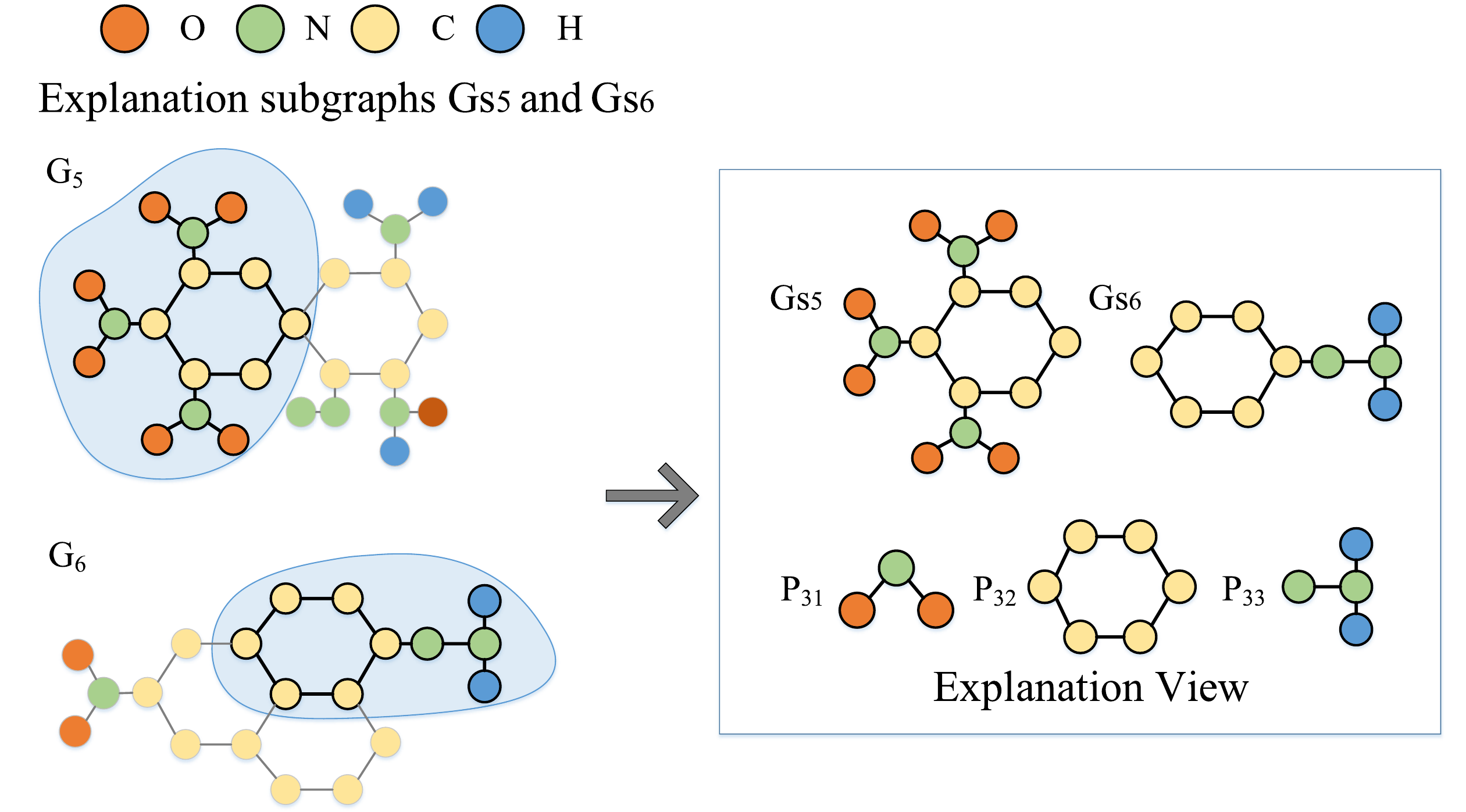}}
\vspace{-2ex}
\caption{An explanation view for a single class label: explanation subgraphs and patterns}

\label{fig-eviews}
\vspace{-3mm}
\end{figure}
\subsection{Explanation Views}
\label{sec:gnn_views}

We now extend graph views as explanation 
structures for \gnn. 
We start with the concept of {\em explanation subgraphs} 
to capture the fraction of a graph that 
is responsible for its label assigned by a \gnn-based classifier $\M$. We then introduce view-based explanation. 

\vspace{-0.6mm}
\stitle{Explanation Subgraphs}.
Given a \gnn $\M$ and 
a single graph $G\in \G$ with label 
$\M(G)$ = $l\in\L$, we say that a 
subgraph of $G$ is 
an {\em explanation subgraph} of 
$G$ for $\M$ {\em \wrt a label $l$}, denoted as 
 $G^l_s$, if 
\begin{itemize}
\item $\M(G)$ = $\M(G^l_s)$ = $l$ (``{\em consistent}''), and
\item  $\M(G\setminus G^l_s)\neq l$ (``{\em counterfactual}''). 
\end{itemize}

Here $G\setminus G^l_s$ is the subgraph 
obtained by removing $G^l_s$ from $G$. 
An explanation subgraph $G^l_s$ of $G$ 
with label $l$ is a subgraph of $G$ that clarifies ``why'' $\M(G)$ = $l$ in terms of counterfactual causality~\cite{verma2021counterfactual}. That is, it is consistently assigned the same label $l$ by $\M$ as $G$, and if ``removed'' from $G$, 
$\M$ assigns the remaining fraction of $G$ a different label
for a graph database $\G$ and a set of 
class labels $\L$,  
one can fine-tune a set of labels of interests 
from $\L$ and generate explanation 
subgraphs accordingly (as will be discussed).  

\vspace{-0.6mm}
\stitle{Explanation Views}. 
Given a graph database $\G$, a \gnn classifier $\M$, 
and a user-specified label of 
interest $l\in\L$, we consider 
a {\em label group} 
$\G^l\subseteq \G$ as the 
set of graphs with assigned label $l$. 
An {\em explanation view} of 
$\G$ for $\M$ \wrt $l$ is a graph view \gvl = $(\P^l, \G^l_s)$,  
    where 
    \begin{itemize}
   \item 
    $\G^l_s$ is a set of explanation subgraphs of the label group $\G^l$, such that 
    for each graph $G\in \G^l$, there is an 
    explanation subgraph\footnote{We remark that $G_s$ may be disconnected; for this case, each  disconnected component can be considered as an explanation subgraph ``corresponding'' to $G$.}$G_s$ of $G$ in $\G^l_s$  
    and 
    \item $\P^l$ is a set of 
    graph patterns, such that 
    the nodes of $\G^l_s$ are covered 
    by the graph patterns in $\P^l$. 
    \end{itemize}
    
Intuitively, an explanation view \gvl
provides a two-tier interpretation 
of \gnns in terms of a specific label $l$ of interest. The ``lower-tier'' explanation subgraph explains \gnn \wrt a label of interest 
with consistent and counterfactual properties. 
The ``higher-tier'' patterns serve as a concise summary to enable easy accessing, querying, and inspection of the classification and explanation results. Prior works verify the need and effectiveness of two-level explanation structures: a higher-level example, global “concept”,  or
“prototype” patterns of each class (similar to our higher-tier patterns) for effective querying and summary of lower level detailed explanations ~\cite{cai2019effects, xuanyuan2023global, dai2022towards}.
\begin{example}
\label{exa-notation}
For a pretrained \gnn model 
$\M$ in 
Example~\ref{exa-motivation}, 
we observe (and experimentally 
verified) the following. 
(1) For the label group mutagen $\{G_1, G_2\}$ classified by the \gnn $\M$, 
an explanation view $\G^{mutagen}_\V$ = 
$(\P^{mutagen}, \G^{mutagen}_s)$.  (a) $\G^{mutagen}_s$ = $\{G_{s1}, G_{s2}\}$
contains two explanation subgraphs 
$G_{s1}$ of $G_1$ and $G_{s2}$ of 
$G_2$. $\M$ will incorrectly classify 
the remaining fraction of 
$G_1$ (resp. $G_2$) obtained by removing 
$G_{s1}$ (resp. $G_{s2}$) as nonmutagen.  
(b) $\P^{mutagen}$ = $\{P_{11},P_{12}\}$ 
contains two graph patterns that concisely summarize the structural information of all 
the explanation subgraphs in 
$\G^{mutagen}_s$, with all the nodes 
covered by $\P^{mutagen}$. 
(2) Similarly, for the label nonmutagens, an 
explanation view $\G^{nonmutagen}_\V$ 
contains explanation subgraph set $\G^{nonmutagen}_s$ as 
$\{G_{s3}, G_{s4}\}$, and a set of
patterns $\P^{nonmutagen}$ 
= $\{P_{21}, P_{22}\}$. 

Consider adding two more graphs $\{G_5, G_6\}$  in Figure~\ref{fig-eviews} to mutagen group classified by $\M$. An explanation view 
of $\{G_5, G_6\}$ for 
$\M$ with label mutagen is illustrated on the 
right side, which contains two 
new explanation subgraphs $G_{s5}$ 
and $G_{s6}$, and a pattern set 
$\{P_{31},P_{32},P_{33}\}$. 
Ideally, one wants to efficiently  
maintain the explanation view 
$\G^{mutagen}_\V$ by properly enlarging $\P^{mutagen}$ and $\G^{mutagen}_s$ 
only when necessary. For example, it suffices 
to keep only $P_{11}$ or 
$P_{32}$, and $P_{12}$ or $P_{31}$, 
in $\P^{mutagen}$. 
\end{example}

Given a set of interested labels\footnote{We abuse 
the notation $\L$ and let it denote a set of user's interested labels.} 
$\L$, and graph database $\G$ where 
each graph $G\in \G$ is assigned one 
of the labels $l\in \L$, we are interested in generating and maintaining 
a set of $|\L|$ explanation views 
$\G^{\L}_{\V}$ = $\{\G^l_\V|l\in \L\}$, 
one for each label group. 
Note that a label group may have multiple potential explanation views. We will elucidate the quality measures to determine the optimal explanation views for a given label group, 
and introduce algorithms to 
compute and maintain explanation views 
in the following sections.

\section{View-based Explanation}
\label{sec-view}

Given a graph database $\G$ and a 
\gnn $\M$, there naturally 
exist multiple explanation 
views for the classification results of $\M$ over $\G$. {\em How to 
measure their ``goodness''}? 
We start with desirable properties 
and introduce quality measurements (\S\ref{sec-quality}), followed by the problem formulation (\S\ref{sec-prob}) and an analysis of 
properties and complexity (\S\ref{sec-hard}). 
\vspace{-1.5ex}
\subsection{Quality Measures}
\label{sec-quality}
\vspace{-1.5ex}
\stitle{Explainability}. Our 
first measure quantifies how 
well the ``lower tier'' explanation subgraphs of explanation views 
interpret a \gnn $\M$, naturally under the {\em influence maximization} 
principle: An explanation view has better 
explainability if its 
explanation subgraphs involve 
more nodes with features that can  
maximize their influence via a random walk-based message passing process (following 
Eq.~\ref{eq-prop}).  
For \gnns that learn and infer via 
feature propagation, this principle  
has been consistently adopted to understand the accuracy of \gnns~\cite{xu2018representation,zhang2021grain} and their robustness~\cite{bajaj2021robust},
\ie the likelihood the labels are changed 
when the features of such nodes are
changed. 

\vspace{.5ex}
Given a label group $\G^l$ = $\{G_1, \ldots, G_n\}$ and $k$-layer GNN $\M$, 
the 
{\em explainability} of an 
explanation view \gvl = $(\P^l, \G^l_s)$  
for $\M$ over $\G^l$ is quantified as:
\begin{equation}
        f(\G^l_\V) = \sum_{G_{si}\in\G^l_s} \frac{I(V_{si}) + \gamma  D(V_{si})}{|V_i|} 
        \label{eq:goodness}
\end{equation}
where (1) $V_{si}$ is the 
node set of an explanation subgraph 
$G_{si}$ of $G$ ($G_{si}\in \G^l_s$, and 
$G_i\in \G$ for $i\in[1,n]$), and 
$V_i$ is the node set of $G_i$ ($V_{si}\subseteq V_i$); 
(2) $I(V_{si})$ is a 
function that quantifies the ``influence'' of 
the features of the node set $\V_s$ via feature propagation in the inference process of $\M$,  and (3) 
$D(V_{si})$ is a diversity measure to 
capture influence maximization. Here a weight $\gamma\in[0,1]$ is introduced 
to balance between feature influence 
and diversity. 
    
We next introduce the two functions $I(\cdot)$ and $D(\cdot)$. 

\eetitle{Feature Influence}. Following feature sensitivity and influence analysis in \gnns~\cite{xu2018representation,zhang2021grain}, we introduce an {\em influence} score. Given a graph $G$ with node set $V$, 
the influence of a node $u$ on another node $v$ at the $k$-layer propagation is defined as the L1-norm of the expected Jacobian matrix~\cite{xu2018representation}:
    \begin{equation}
        I_1(v,u) = \big|\big| \mathbf{E}[\partial X_v^k / \partial X_u^0] \big| \big|_1 
    \end{equation}
Intuitively, $I_1(v,u)$ quantifies how 
``sensitive'' the representation $X^k_v$ of 
a node $v$ at the $k$-th layer of $\M$ is, upon changes of the representation $X^0_u$ 
at the input layer of $\M$ for a given connected 
node $u$; in other words, how 
``influential'' $u$ to $v$ is via 
feature propagation. 

Given a targeted node $v$, the influence score 
of a node $u\in V$ to $v$ can be normalized as 
    \begin{equation}
        I_2(u, v) = \frac{I_1(v,u)}{\mathop{\sum}_{w \in V} I_1(v,w)}
        \label{eq:norm_inf}
    \end{equation}

Given a threshold $\theta$ and a set of nodes 
$V_s\subseteq V$, we say that a node $v$ is 
    {\em influenced} by $V_s$ if 
    there exists a node 
    $u\in V_s$, such that $I_2(u, v)\geq \theta$. 
    The influence score 
    of $V_s$, denoted as $I(V_s)$, is in turn defined as the size of 
    nodes influenced by $V_s$, 
    \ie
    \begin{equation}
        I(V_s) = \left|\{v|I_2(u, v)\geq \theta, u\in V_s, v\in V\}\right|
        \label{eq:influece-score}
    \end{equation}

\eetitle{Neighborhood Diversity}. The second function aggregates a diversity measure among 
the neighboring nodes influenced by the explanation 
subgraphs via feature propagation. 
Recall that the node representation  
of $v$ at the output layer 
(the $k$-th layer) of the \gnn $\M$ is 
$X^k_v$. Let $r(v,d)$ be the set of 
nodes in $V$ such that for each node $v'\in r(v,d)$, the distance $d(X_v^k,X_{v'}^k)$ between nodes 
$v$ and $v'$ is bounded by a threshold $r$, \ie 
$r(v, d)$ = $\{v'|d(X_v^k,X_{v'}^k)\leq r, v, v'\in V\}$. 

The function $D(V_s)$ quantifies a {\em neighborhood diversity} as size of 
the union of $r(u,d)$ for each node $u$ 
influenced by $V_s$, 
\ie 
\begin{small}
    \begin{equation}
        D(V_s) = \left|\mathop{\bigcup}_{v: I_2(V_s, u, v)\geq \theta}r(v,d)\right|
        \label{eq:diversity-score}
    \end{equation}
\end{small} 

Here the distance function $d(\cdot)$ can be any embedding 
distance measure, such as the normalized Euclidean distance.

Putting these together, an explanation view 
with higher explainability favors 
explanation subgraphs that 
{\bf (1)} have greater feature influence following feature propagation process, and 
{\bf (2)} influence more nodes with larger neighborhood diversity.  

\stitle{Coverage}. 
Besides ``lower-tier'' explainability, 
we also expect the ``higher-tier'' patterns 
of an explanation view
to {\em cover} a desirable amount of 
nodes for each label group of interests. Better still, 
such coverage constraints should be 
explicitly configurable by users. These coverage constraints become especially valuable when conducting label-specific analyses on multiple labeled groups~\cite{lin2020identifying, zhang2021automatic}.

Given a label group $\G^l$ = 
$\{G_1, \ldots G_n\}$, and 
its node set $\V^l$ = $\bigcup_{G_i\in\G^l} V_i$, a coverage constraint 
is a range $[b_l, u_l]$, 
where $0\leq b_l\leq u_l\leq |\V^l|$.
We say that an explanation view 
\gvl = $(\P^l, \G^l_s)$
{\em properly covers} 
the label group $\G^l$ if 
the explanation subgraphs
$\G^l_s$ contain in 
total $n$ nodes from $\V^l$ 
where $n\in [b_l, u_l]$. 
Note that by definition of 
graph views, 
$\P^l$ also covers 
all the nodes from $\G^l_s$. 

\subsection{Explanation View Generation Problem}
\label{sec-prob}
\vspace{-1mm}

\stitle{Configuration}. A {\em configuration}  
$\C$ specifies the following parameters: 
(1) a pair of thresholds $(\theta, r)$ to determine the influence and 
diversity scores in explainability measure; and
(2) a set of coverage constraints $\{[b_l, u_l]\}$ 
for each class label $l\in \L$. 

\vspace{.5ex}
We now formulate the problem of {\em Explanation View Generation}. 

\begin{problem}
 
Given a graph database $\G$, a set of interested labels $\L$ s.t. $|\L|=t$, 
a \gnn $\M$, and a configuration $\C$,
the {\em explanation view generation problem}, 
denoted as \evg, 
is to compute a set of graph views $\G_\V$ = $\{\G^{l_1}_\V, \ldots \G^{l_t}_\V\}$, such that 
$(i\in [1,t])$: 
\begin{itemize}
\item Each graph view
$\G^{l_i}_\V$ = $(\P^{l_i},\G^{l_i}_s)\in \G_\V$ is an explanation view 
of $\G$ for $\M$ \wrt $l_i\in\L$; 
\item Each $\G^{l_i}_\V$ properly covers the label group $\G^{l_i}$; and 
\item $\G_\V$ maximizes an aggregated 
explainability, \ie 
\begin{equation}
\G_\V = 
\argmax\sum_{\G^{l_i}_\V \in \G_\V} f(\G^{l_i}_{\V})
\end{equation}
\end{itemize}
\end{problem}

That is,  we are interested in generating a set of explanation views which maximizes the explainability and simultaneously properly covers $\G$ 
\wrt the configuration for each labeled group. 

\subsection{Hardness and Properties}
\label{sec-hard}

To understand the hardness and feasibility 
of generating explanation views for \gnns, we 
study several fundamental 
issues and properties. 
Our results are established 
for a {\em fixed} \gnn model $\M$. 
We follow the convention in cost analysis 
of \gnns~\cite{gunnemann2022graph}, and say 
that $\M$ is ``fixed'' if it is given, pretrained (thus, its architecture 
and weights no longer change), and incurs an inference cost in polynomial time (\PTIME). 

\stitle{View 
Verification}. 
To understand the hardness of \evg, 
we first investigate 
a ``building block'' decision problem, 
notably, {\em view verification}.  Given $\G$, $\C$, $\L$, a fixed \gnn $\M$, and a two-tier 
structure $\G_\V$ = $(\P, \G_s)$ 
with a pattern set $\P$ and
a set of subgraphs $\G_s$ 
of the graphs from $\G$, it  
verifies if $\G_\V$  
satisfies three constraints simultaneously:  
(\textbf{C1}): it is a graph view of $\G$, 
(\textbf{C2}): if so, if it is an explanation view of the label group $\G^l$ = $\{G\}$, 
where $\M(G)$ = $l$; 
and (\textbf{C3}): if so, if
it properly covers $\G^l$ 
under the coverage constraint in $\C$. 

The hardness of verification 
provides the lower bound 
results for \evg, and its solution 
will be used as a 
primitive operator in \gvex view generation framework (see \S\ref{sec-approx}). And then
we present the following result. 

\begin{lemma}
\label{lm-verifynpc}
Given a graph database $\G$, configuration $\C$, and a two-tier structure $(\P, \G_s)$, the view verification problem is \NP-complete when the \gnn $\M$ is fixed.
\end{lemma}

\vspace{-1mm}
\begin{proofS}
It is not hard to verify that view verification is \NP-hard, given that 
it requires subgraph isomorphism 
tests alone to verify constraint 
\textbf{C1}, which is known 
to be \NP-hard~\cite{floderus2015induced}.  
We next outline an \NP algorithm for the verification problem. 
It performs a three-step 
verification below. 
(1) For \textbf{C1}, it guesses a finite number of matching  
functions in \PTIME 
(for patterns $\P$ and $\G$ with bounded size), and verifies if the patterns induce 
accordingly $\G_s$ via the matching 
functions in \PTIME. If so, 
$\G_\V$ is a graph view. 
(2) To check \textbf{C2}, for 
each graph $G\in \G$ and 
its corresponding 
subgraphs $G_s\in\G_s$, 
it applies $\M$ to verify 
if $\M(G_s)$ = $l$ and 
$\M(G\setminus G_s)\neq l$. 
If so, $\G_\V$ is an 
explanation view for $\G$. 
For a fixed \gnn $\M$, 
it takes \PTIME to do the 
inference. 
(3) It takes \PTIME to 
verify the coverage given 
that subgraph isomorphism 
tests have been performed 
in steps (1) and (2). 
These verify the upper bound 
of view verification. 
\end{proofS}

\vspace{-1ex}
\stitle{Hardness of EVG}. 
Given a threshold $h$, 
the decision problem of \evg is to determine 
if there exists a set of explanation views 
$\G_\V$ for \gnn $\M$ 
with explainability at least $h$ 
under the constraints in $\C$. We present a hardness result below for \evg.  

\begin{theorem}
\label{thm-hardness}
For a fixed \gnn $\M$,  
\evg is (1) $\Sigma^2_P$-complete, and 
(2) remains $NP$-hard even 
when $\G$ has no edges. 
\label{thm-hardness-subgraph}
\end{theorem}

Here a problem is in $\Sigma^2_P$ if an \NP algorithm exists to solve it with an \NP oracle.  
Theorem~\ref{thm-hardness} verifies 
that it is beyond NP to generate explanation views under coverage 
constraints, thus the general \evg problem is hard even for fixed \gnns. 
We outline the hardness analysis below and present the detailed proof in the appendix.  

\begin{proofS} 
(1) \evg is solvable in $\Sigma^2_P$ since we can devise an \NP oracle for view verification by guessesing a set of two-tier view structures $\G_\V$
= $\{(\P,\G_s)_i\}$ ($i\in[1,|\L|]$) and calling the \NP algorithm in the proof of Lemma~\ref{lm-verifynpc} $O(|\L||\P||\G|)$ times to check the fulfillment of the three 
constraints. If so, it then computes $f(\G^l_\V)$ and checks if 
$f(\G^l_\V)\geq h$ in 
\PTIME. 
(2) To see that \evg is $\Sigma^2_P$-hard, 
we construct a reduction from graph satisfiability, a known $\Sigma^2_P$-complete problem~\cite{schaefer2002completeness}. 
(3) To see Theorem~\ref{thm-hardness}(2), we consider 
a special case of \evg. Let $\G$ contains two single graphs $G_1$ and $G_2$, 
each has no edge. For such a case, we prove that 
\evg remains to be \NP-hard through a reduction 
from the red-blue set cover problem~\cite{robert00}, which is \NP-complete. 
This verifies the hardness of \evg for identifying
explanation with coverage requirement alone, as in such case, subgraph isomorphism test is no longer intractable. 
\end{proofS} 

While Theorem~\ref{thm-hardness} tells us that \evg is in general hard, we next 
show its properties that indicate 
feasible algorithms with provable 
quality guarantees in practice. 
 
\stitle{Monotone Submodularity}. 
We first show that the explainability  
$f(\G_\V)$ of an explanation view 
$\G_\V$ is essentially a {\em monotone submodular} set function~\cite{calinescu2011maximizing},  
determined by the nodes from 
its explanation subgraphs. 
Clearly, $f(\G_\V)$ is non-negative.
Denote the set of nodes 
from $\G^l_\V$ as $V_s$, with $V_{s}$ ranges over the node set of $G_{s}$ in $\G^l_s$. 

\begin{lemma}
\label{lm-submodular}
  Given $\G$, $\L$, $\C$, and a fixed \gnn $\M$,  $f(\G_\V)$ is a monotone submodular function. 
  \vspace{-1ex}
\end{lemma}

\begin{proofS}
    We first show that the monotonicity and submodularity of $f(\cdot)$ depend on the two components $I(\cdot)$ and $D(\cdot)$. Then we show that enlarging the node set will never downgrade the feature influence, thus  $I(\cdot)$ is monotonic. Next we systematically discuss the marginal gain of $I(\cdot)$ for any set $V_{s''}\subseteq V_{s'}$ and any node $u\not\in V_{s'}$ under several cases, leading to a conclusion that $|\kw{Inf}(V_{s''} \cup \{u\})| - |\kw{Inf}(V_{s''})| \ge |\kw{Inf}(V_{s'} \cup \{u\})| - |\kw{Inf}(V_{s'})|$. Finally, we show that the similar properties of $D(\cdot)$ can be analyzed in the same manner. The complete proof is in the appendix. 
\end{proofS}

We next present an algorithm framework, 
denoted as \textbf{GVEX}, to 
solve the \evg problem. We show that there exists feasible approximations for \evg in \S\ref{sec-approx}, and then introduce an efficient algorithm to maintain explanation views in \S\ref{sec-stream}. 

\section{Generating Explanation Views}
\label{sec-approx}

Our main results below show that there exist feasible algorithms to generate 
explanation views with 
guarantees on both explainability and coverage 
constraints, for \gnn-based graph classification.

\begin{theorem}
\label{thm-approx}
Given a configuration $\C$, graph database $\G$, and a $k$-layer \gnn $\M$ over label set $\L$,  there is a $\frac{1}{2}$-approximate algorithm for generating explanation views, and takes $O(
|\G|
|V_m|^3 + 
|\G|
|V_m||\L|k\cdot\C.u_l(dD+D^2)+ N(N+T))$ time.
\end{theorem}

Here, $|\G|$ is the number of graphs in $\G$; $V_m$ refers to the largest node set of 
a graph in $\G$, $d$ and $D$ are the 
average degree and the number of features per node, 
and $N$ and $T$ are the total number of 
verified patterns and the cost for 
single isomorphism test, respectively. 

We start by 
presenting an approximation algorithm 
that generates an explanation view for 
a single label $l\in\L$ and a single 
graph $G\in\G^l$.  Our general approximation scheme  calls this algorithm for each graph $G\in \G^l$ to assemble an explanation view 
$\G^l_\V$, and then returns 
a set of explanation views $\G_\V$ as 
$\bigcup_{l\in\L}\G^l_\V$.

\stitle{``Explain-and-Summarize''}. 
Our general idea is to follow a two-step  
``explain-and-summarize'' strategy. 
(1) In the ``explain'' stage, the algorithm 
selects high-quality nodes to induce 
``lower-tier'' explanation subgraphs for $\M$
that can maximize the explainability score, and 
meanwhile, ensures the coverage 
constraints in the configuration $\C$. 
(2) The ``summarize'' stage produces, 
as ``higher-tier'' structure, 
a set of graph patterns 
that ensures to cover the nodes of 
the explanation subgraphs. The computed 
components are then assembled to yield 
the desired explanation views. The output of the two stages captures explainability and provides queryable property, respectively.

To ensure the quality guarantee and 
efficiency, the algorithm adopts several primitive operators, 
which are described below. 

\eetitle{Verifiers}. 
The verifiers are efficient 
operators that implement the view verification to check the constraints 
\textbf{C1}-\textbf{C3} as specified by 
 configuration $\C$ (see the proof 
of Lemma~\ref{lm-verifynpc}), whenever 
a two-tier structure $(\P, \G_s)$ 
is in place. 
\gvex calls two primitive 
verifiers: 
\begin{itemize}
\item
a \gnn inference operator \gnninf, which efficiently infers the label of a subgraph $G_s$ of $G$ and its counterpart 
$G\setminus G_s$ with $\M$ (constraint \textbf{C2}); and 
\item a pattern matching operator \pmatch, that performs fast node-induced subgraph isomorphism and 
checks whether the nodes 
in explanation subgraphs are covered by patterns (constraint \textbf{C1}), and 
are also properly covered (constraint \textbf{C3}).  
\end{itemize}
Since the view verification problem is \NP-complete, we approach it by addressing constraints using two efficient primitive verifiers. It allows us to offer a feasible solution and establishes lower bound results for \evg. In practice, they  
can be supported by invoking established 
solutions, \eg parallel \gnn inference~\cite{gao2022efficient,kaler2022accelerating} and subgraph pattern matching~\cite{shang2008taming,han2013turboiso}, 
respectively.

\setlength{\textfloatsep}{0pt} %
\begin{algorithm}[htb!]
    \renewcommand{\algorithmicrequire}{\textbf{Input:}}
    \renewcommand{\algorithmicensure}{\textbf{Output:}}
    \caption{Algorithm~\gvapprox (for a single graph $G$)}
    \begin{algorithmic}[1]
        \REQUIRE A graph $G$, a GNN $\M$, a 
        label $l$, a configuration $\C$;  
        \ENSURE an explanation view $\G_{\V}^l$ for $G$ and $l$.  
        \STATE set $V_S := \emptyset$; set $V_u := \emptyset$; set $\G_s^l := \emptyset$; set $\G_{\V}^l$ := $\emptyset$; set $\P$:=$\emptyset$; \\
        \STATE Invoke \gnninf to precompute Jacobian Matrix $M_I$ of $G$;  \\
        /* {\em explanation phase */} \\
        \WHILE{$|V_S| \textless \C.u_l$ \And $V \backslash V_S\neq\emptyset$}
        \FOR {$v \in V \backslash V_S$}
            \IF {\vpexp($v$,$V_S$,$G$,$G^l_s$, $\C$,$\M$)} 
                \STATE $V_u := V_u \cup \{v\}$; 
            \ENDIF
        \ENDFOR
        \STATE $v^* := \mathop{\mathrm{argmax}}_{v' \in V_u} (f(V_S \cup v')-f(V_S))$; 
        \STATE $V_S := V_S \cup \{v^*\}$;
        \STATE extend $\G_s^l$ with the selected node $v^*$; 
        \ENDWHILE
        /* {\em use candidate set $V_u$ to satisfy lower bound requirement} */ \\
        \WHILE {$|V_S|\textless\C.b_l$ \And $V_u\neq\emptyset$}
        \FOR{$v' \in V_u$}
             \IF{\vpexp($v'$,$V_S$,$G$,$G^l_s$, $\C$,$\M$)}
             \STATE $v^* := \mathop{\mathrm{argmax}}_{v' \in V_u} (f(V_S \cup v')-f(V_S))$; \\
            \STATE $V_S := V_S \cup \{v^*\}$; \\
            \STATE extend $\G_s^l$ with the selected node $v^*$; \\
            \ENDIF
             \ENDFOR 
             /* {\em no ``large enough'' explanation that satisfy lower bound} */\\
             \IF{$V_u = \emptyset$ \And $|V_S|\textless\C.b_l$} 
            \RETURN $\emptyset$; 
            \ENDIF
        \ENDWHILE
        
        /* {\em summary phase */} \\
        \STATE  $\G_{\V}^l$ := \psum$(\G_s^l, V_S)$; 
        \RETURN  $\G_{\V}^l$; 
    \end{algorithmic}
  \label{alg:node_selection}
\end{algorithm}
\setlength{\textfloatsep}{12pt plus 2pt minus 2pt}

\eetitle{Pattern generators}. 
We use a second operator \pgen 
to extract a set of 
pattern candidates to be 
verified by \pmatch, from a set of 
explanation subgraphs. 
The operator exploits 
minimum description length (MDL) 
principle and 
conducts a constrained 
graph pattern mining process. 
It can be implemented by invoking 
scalable pattern mining 
algorithms, \eg~\cite{yan2002gspan}. Advanced mining algorithms developed in the future can be used to enhance the \pgen method further.

\stitle{Algorithm}. The algorithm, denoted as \gvapprox (Algorithm \ref{alg:node_selection}), computes an explanation 
view $\G_{\V}^l$ for a label $l\in \L$ and 
a graph $G$.  

\eetitle{Initialization} (lines~1-2). \gvapprox 
initializes and maintains the following 
auxiliary structures (i.e., initialized globally once, and not re-initialized for each graph): (1) two node-sets $V_u$ 
and $V_S$, to store the candidate nodes 
and the selected ones that contribute to inducing explanation subgraphs, respectively; 
(2) a set $\G_s^l$
of explanation subgraphs to be summarized, 
and the explanation view $\G_{\V}^l$. 
In addition, it also pre-computes 
the Jacobian matrix $M_I$ with 
the operator \gnninf. Note that 
this once-for-all inference also 
prepares node representations 
that are needed 
to compute 
$I(\cdot)$ and $D(\cdot)$. 

\eetitle{Explanation phase} (lines 3-10). 
In this phase, \gvapprox dynamically 
expands a set of selected nodes $V_S$ with 
high influence scores to construct explanation 
subgraphs. (1) It first checks if 
a new node in $\V\setminus V_S$ 
can contribute to ``extend'' an existing 
explanation subgraph in 
its original graph $G\in \G$, 
by invoking procedure~\vpexp 
(line~7, to be discussed).  
(2) Upon the enlargement of $V_u$, it adopts a 
greedy selection strategy to 
iteratively choose the node $v^*$ from $V_u$
that can maximize the 
marginal gain (line 9).
It then extends $V_S$ with $v^*$, 
and updates the current set of 
explanation subgraphs by 
including $v^*$ and its induced 
edges in $G$ until finish condition (line~3).  
\gvapprox then takes care of the 
lower bound requirement (lines~10-17). 
If $\G^l_s$ contains too 
few nodes $V_S$ to 
satisfy the lower bound 
requirement $\C.b_l$ (line~10), 
it repeats the greedy 
selection from 
the candidate set $V_u$, 
until $\G^l_s$ grows to 
desired size or 
no candidate node is 
available (line 10). 
If all candidates are 
processed and $\G^l_s$ 
is still small, 
\gvapprox returns 
$\emptyset$ (lines~16-17). 

\eetitle{Summary phase} (line 18). In this phase, 
\gvapprox invokes procedure~\psum to construct 
patterns that cover $V_S$ with a 
small number of patterns. It then constructs and returns $\G_\V^l$. 

We next present the details of the procedures
\vpexp and \psum. 

\setlength{\textfloatsep}{0pt}
\floatname{algorithm}{Procedure}
\begin{algorithm}[htb!]
    \renewcommand{\algorithmicrequire}{\textbf{Input:}}
    \renewcommand{\algorithmicensure}{\textbf{Output:}}
    \caption{Procedure \vpexp($v$, $V_S$,$G$,$G_s$,$\C$,$\M$)} 
    \begin{algorithmic}[1]
        \STATE Update explanation subgraph $G_s$ with $v$ to $G_t$; \\
        /*{\em invokes \gnninf to verify 
        constraint \textbf{C2} (``View verification''; \S~\ref{sec-hard})}*/ \\
        \IF {$\M(G_t) \neq \M(G)$ \OR $\M(G \backslash G_t) = \M(G)$}
            \RETURN \False
        \ENDIF
        \STATE set $V_t := V_S \cup \{v\}$; \\
            \IF {$|V_t|\ge \C.u_l$} 
                \RETURN \False; 
            \ENDIF
        \RETURN \True;    
    \end{algorithmic}
    \label{procedure:extendable}
\end{algorithm}
\setlength{\textfloatsep}{12pt plus 2pt minus 2pt}
\vspace{-4mm}

\stitle{Procedure~\vpexp}. 
The procedure~\vpexp implements the 
view verification algorithm (see \S\ref{sec-view}). It
invokes 
the two verifier operators to 
determine if the explanation 
subgraphs, in particular, the 
nodes $V_S$ that are used to 
induce them, can be ``extended''. 
It first constructs an 
explanation subgraph by 
augmenting the current fraction 
that belongs to $\G_s^l$ with 
the node $v$ to be verified, 
and follows the verification process 
to check the 
invariant conditions, 
\ie consistency, counterfactual 
explanation, and coverage conditions. 

\begin{figure}[tb!]
\vspace{-1ex}
\centering
\centerline{\includegraphics[width=0.92 \linewidth]{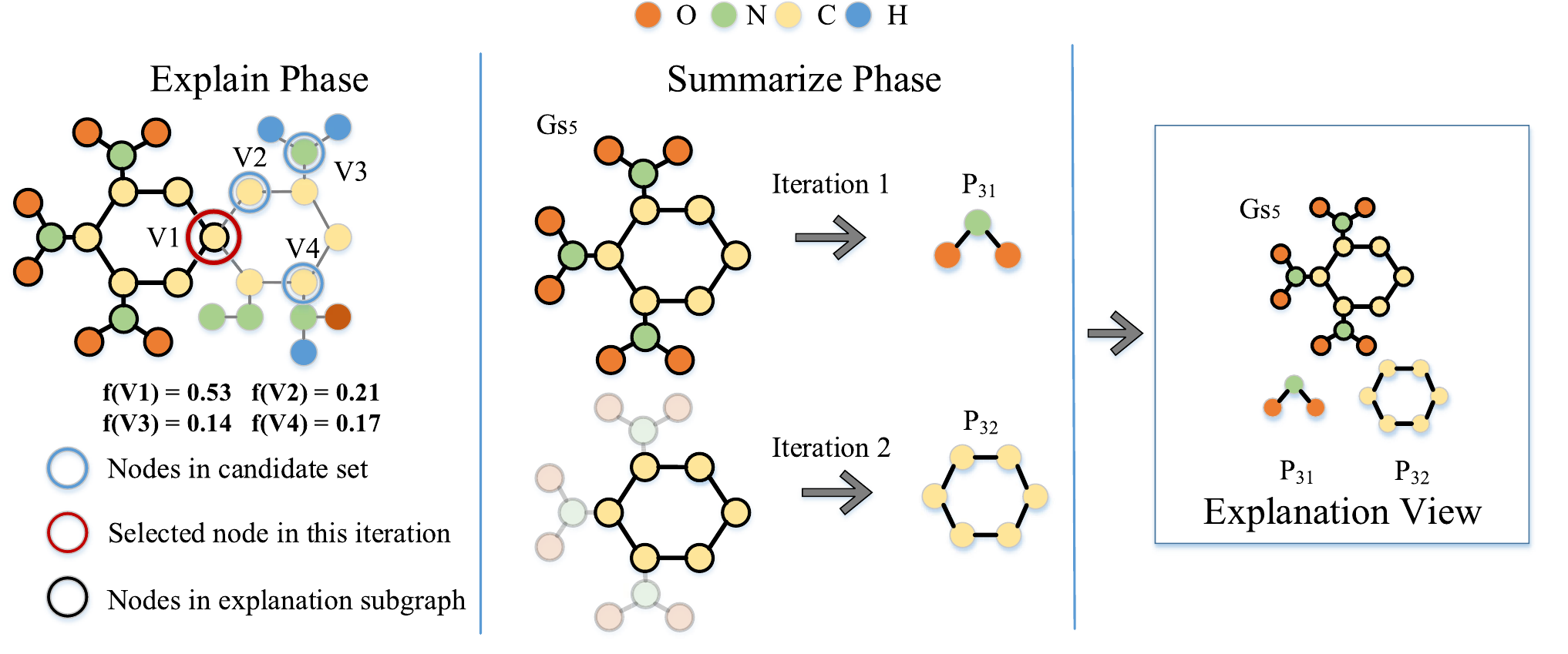}}
\vspace{-3mm}
\caption{``Explain-and-Summarize'': an illustration}
 \vspace{-1ex}
\label{fig-alg-approx}
\end{figure}

\begin{example}
\label{exa-explain}
Figure \ref{fig-alg-approx} illustrates the "Explain-and-Summarize" process, 
with a configuration $\C$ = 
$(0.14, 2, (0, 15))$.  
(1) In the explanation phase, 
\gvapprox identifies four candidate nodes 
$\{v_1$, $v_2$, $v_3$, $v_4\}$, which pass the
verification in \vpexp. These nodes are stored in 
$V_u$. 
(2) It then greedily selects 
the node $v_1$ with the highest gain 
on explainability for 
$G^l_s$ (with a score 
$0.53$ in our experiment). 
This repeats until $V_u$
are processed and an explanation 
subgraph is induced as $G_{s5}$. Since the upper bound $u_l$ = $15$ is reached, $G_{s5}$ is returned as an explanation subgraph. 
\end{example}

 \vspace{-1ex}
\stitle{Procedure~\psum}. Given the explanation subgraphs $\G_s^l$ 
induced from explanation phase, procedure \psum 
computes a set of ``higher-tier'' 
patterns $\P$ to cover the nodes of 
$\G_s^l$. Meanwhile, it is 
desirable for $\P$ to cover the edge set of $\G_s^l$ as much as possible. 
Given a pattern $P\in\P$ and graphs 
$\G^l_s$ with node set $V_S$ and edge set $E_S$, we denote the nodes and edges in $\G^l_s$
it covers as $P_{V_S}$ and $P_{E_S}$, respectively. 
Let each $P$ be ``penalized'' by a normalized weight (as the Jaccard distance) between 
$E_S$ and $P_{E_S}$, \ie 
$w(P)$ = 1- $\frac{|P_{E_S}|}{|E_S|}$ 
(note $P_{E_S}\subseteq E_S$). 
The above requirements can be further
formulated as an optimization 
problem: 
\begin{itemize}
\item \textbf{Input}: explanation subgraphs $\G_s^l$; 
\item \textbf{Output}: a pattern set $\P^l$, 
such that (1) $\bigcup_{P\in\P^l}$($P_{V_S}$) = $V_S$ and (2) $\P^l$ = 
$\arg\min\sum_{P\in\P^l} w(P)$. 
\end{itemize}

The procedure \psum solves the 
above problem by conducting a constrained pattern mining 
on explanation subgraphs $\G_s^l$. 
It invokes operator \pgen to iteratively generate a 
set of pattern candidates (line~3), 
and subsequently adopts a greedy strategy to dynamically select  a pattern $P^*$ that maximizes a gain 
ascertained by covered nodes $\P^*_{V_S}$ in $V_S$ 
with the smallest weight. $\P^l$ is enlarged with $\P^*$ accordingly.  
Post the selection of the currently optimal patterns, the matched nodes in $V_S$ are reduced; 
and the weights of the patterns are updated 
accordingly. This allows us to gradually 
acquire the final explanation view and reduce the 
edges ``missed'' by $\P^l$.  

\begin{lemma}
\label{lm-coverage}
For a given set of explanation subgraphs $\G^l_s$, 
procedure~\psum is an $H_{u_l}$-approximation of 
optimal $\P^l$ that ensures 
node coverage (hence satisfies coverage constraint 
in $\C$). Here, $H_{u_l}$ = $\sum_{i\in[1,\C.u_l]}\frac{1}{i}$ is the $u_l$-th Harmonic number ($\C.u_l\geq 1$).
\end{lemma}

The quality guarantee can be verified by 
performing an approximate preserving reduction 
from the optimization problem to the minimum weighted 
set cover problem, for which a greedy selection 
strategy ensures an $H_d$-approximation with 
$d$ the largest subset size~\cite{ajami2019enumerating}, 
which is in turn bounded by $\C.u_l$ for the patterns 
over node-induced subgraphs $\G^l_s$. 
We present the detailed analysis in the appendix. 

\begin{example}
\label{exa-summary}
Continuing Example~\ref{exa-explain}, given 
$G_{s_5}$, \psum 
generates a small set of $2$ pattern candidates: $P_{31}$ and $P_{32}$.  It finds that $P_{31}$ covers 9 nodes in $G_{s5}$, and specifically, capturing the presence of three nitro groups. Consequently, $P_{31}$ is selected as the best pattern. 
The nodes already covered by $P_{31}$ are masked, and a next pattern, $P_{32}$ (carbon ring), is chosen as the second pattern that 
further covers 6 nodes in the remaining part. The 
two patterns properly cover all nodes of $G_{s5}$, 
with a small number ($3$) of edges 
uncovered. By incorporating $G_{s5}$ with a pattern set 
$\P$ = $\{P_{31}, P_{32}\}$, an 
explaination view $\G^l_\V$ is 
constructed as 
$(\P, G_{s5})$. 
\end{example}

\vspace{-1mm}
\stitle{Correctness \& Approximability}.
Algorithm \gvapprox terminates when: 
all nodes in $V$ are processed 
($V\setminus V_S$ = $\emptyset$), or 
all candidates in $V_u$ are exhausted ($V_u$ is $\emptyset$). 
When it terminates with a non-empty $G^l_s$, 
it correctly ensures that $G^l_s$ is an
explanation subgraph, as guarded by the verification 
of the three constraints \textbf{C1}-\textbf{C3} 
in view verification (by invoking procedures~\vpexp and~\psum). 

To see the approximation guarantee, observe that \gvapprox 
generates $\G^l_s$ by carefully constructing $V_S$ that satisfies the coverage constraint ($|V_S|\in[b_l, u_l]$). Given Lemma~\ref{lm-submodular}, it essentially solves \evg as a monotone submodular maximization problem under 
a range cardinality constraint. This allows us to reduce \evg to 
a fair submodular maximization problem~\cite{halabi2020fairness}. 
The latter chooses a node set that maximizes a monotone 
submodular function under ranged coverage constraint (which is set as ($[\C.b_l,\C.u_l]$ in our case). The $\frac{1}{2}$-approximation~\cite{halabi2020fairness} 
carries over for \evg, as \gvapprox carefully selects 
nodes with two 
invariants: (1) whenever $|V_S|\leq \C.u_l$, it  
improves explainability of $\G^l_s$ via greedy strategy that ensures $\frac{1}{2}$-approximation by submodular maximization under metroid constraints~\cite{chakrabarti2015submodular}, 
and (2) if $|V_S|\leq \C.b_l$, 
it continues enlarging $\G^l_s$ with $V_u$ that 
gathers ``back up'' nodes, this does not hurt 
the guarantee on explainability due to its {\em monotonic 
non-decreasing} property.

\floatname{algorithm}{Algorithm}
\begin{algorithm}[tb!]
    \renewcommand{\algorithmicrequire}{\textbf{Input:}}
    \renewcommand{\algorithmicensure}{\textbf{Output:}}
    \caption{Algorithm~\gvstream (for a single graph $G$)}
    \begin{algorithmic}[1]
        \REQUIRE a graph $G$ with label $l$, a GNN $\M$, a configuration $\C$; 
        \ENSURE An explanation view $\G^l_\V$;
        \STATE set $V_S := \emptyset$; set $\G^l_\V := \emptyset$; set $\P_c := \emptyset$; set $V_u := \emptyset$; 
        \FOR {each arriving node $v \in V$ (as a node stream)}
        \STATE invoke \kw{IncEVerify} to update Jacobian Matrix; 
        \STATE $w(v) := f(V_S \cup \{v\}) -f(V_S)$;

        \STATE $V_u := V_u \cup \{v\}$; 
        \IF {\kw{VpExtend}$(v, V_S,G,G_s,\C,\M)$} 
            \STATE $V_S := $\kw{IncUpdateVS}$(v,V_S,V,G,G_s)$;
        \ENDIF        
        \IF{$v \in V_S$}
            \STATE $\P_c$ := \kw{IncUpdateP}$(v,V_S,\P_c)$; 
        \ENDIF
        \ENDFOR
        \STATE use set $V_u$ to update $V_S$ to satisfy lower bound constraint $\C.b_l$; \\       
        \RETURN $\G^l_s$ as $(\P, \G_s)$; 
    \end{algorithmic}
  \label{alg:streaming_selection}
\end{algorithm}

\stitle{Time Cost}. ~\gvapprox incurs a one-time cost in $O(|V|^3)$
to compute Jacobian matrix (line~2). It takes at most
$|V|$ rounds in generating $V_S$. For a fixed \gnn with 
$k$-layers, a full inference takes $O(k|V_S|(dD+D^2))$~\cite{zhou2021accelerating}, 
where $d$ and $D$ refer to the average degree 
of $G^l_s$ and the number of features per node. 
Thus in each round, \vpexp takes $O(k\cdot\C.u_l(dD+D^2))$ time to verify 
if $G^l_s$ remains to be an explanation subgraph.  
The total time cost of \psum 
is $O(N*T + N^2)$, where $N$ is the number of 
verified patterns (each with at most $\C.u_l$ 
nodes) from $G^l_s$, and 
$T$ is the time cost of \pmatch. 
Hence the total cost is $O(|V|^3 + k\cdot\C.u_l(dD+D^2)+ N(N+T))$. 
In practice, $d$ and $D$ are small, and
 $N$ and $T$ are also small due to 
bounded pattern and graph size. 

To generate $\G_\V$ over 
$\G$ and $\L$, one invokes 
\gvapprox at most $|\G|$ 
times. The overall 
time cost is: 
$O(|\G||V_m|^3 + |\G||V_m||\L|k\cdot\C.u_l(dD+D^2)+ N(N+T))$, with $V_m$ the largest node set of 
a graph in $\G$, and the rest terms 
scale to their counterparts for $\G$.

\section{Fast Streaming-based Algorithm}
\label{sec-stream}

Algorithm~\gvapprox requires the generation of 
all explanation subgraphs to complete the 
generation of explanation views. As such, \gnn 
inference or pattern generation alone can be major bottlenecks when $G$ is large. 
Moreover, users may also want to interrupt view generation to 
investigate and ad-hocly query for specific explanation structures.  
In response, we next outline an algorithm to 
{\em incrementally} maintain explanation 
views as it scans over $G$ as a stream of nodes. 

\vspace{-1ex}
\begin{theorem}
\label{thm-online}
Given a configuration $\C$, graph database $\G$, 
\gnn $\M$,  there is an online algorithm that maintains 
explanation views with a $\frac{1}{4}$-approximation. 
\end{theorem}

The above approximation ratio holds for 
an optimal explanation view on the ``seen'' 
fraction of $\G$, thus is a weaker 
form of guarantee; yet this provides a pragmatic 
solution for large $\G$.

\begin{figure}[tb!]
\vspace{-1ex}
\centering
\centerline{\includegraphics[width=0.55 \linewidth]{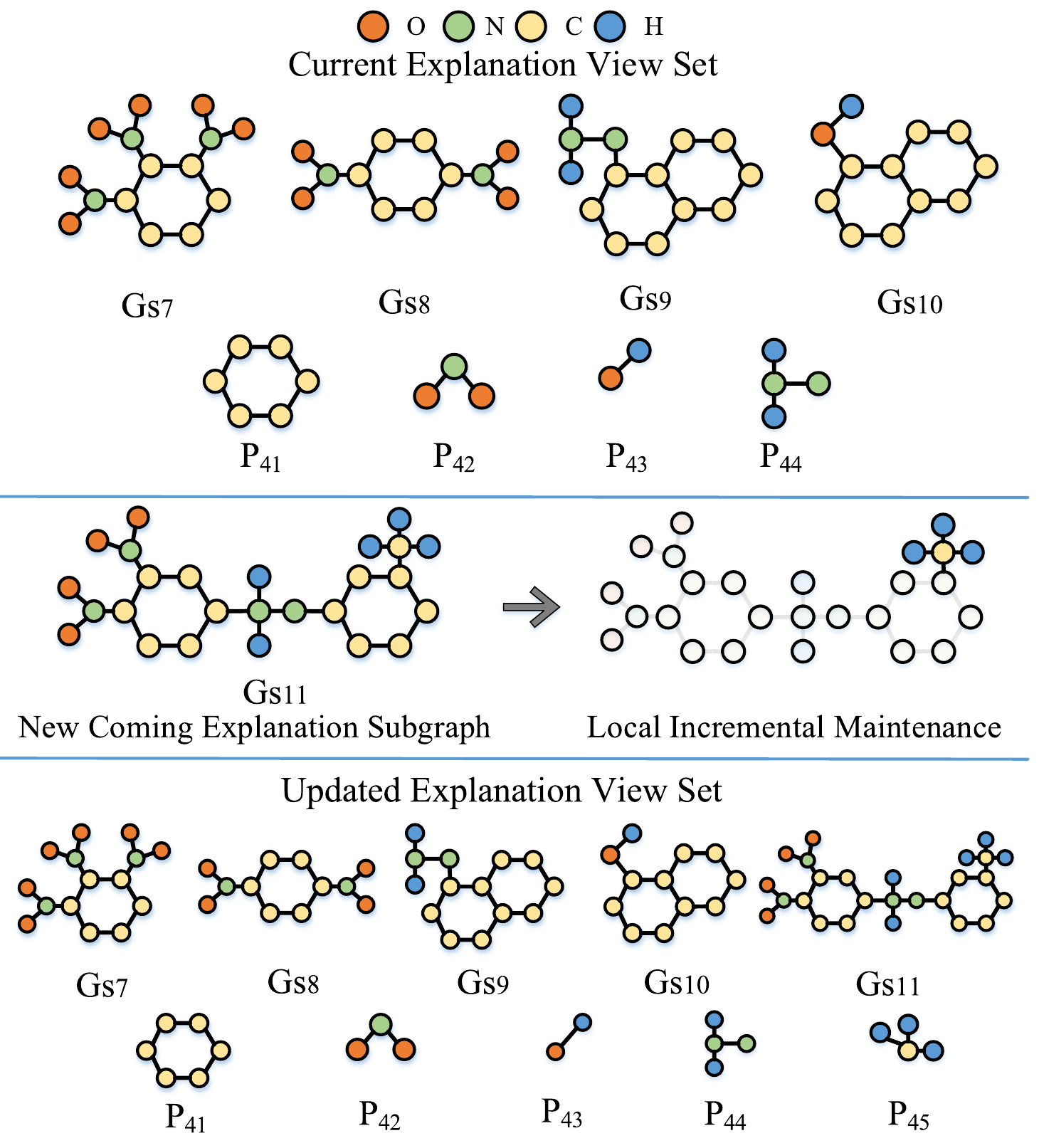}}
\vspace{-3mm}
\caption{Incremental generation of explanation views}
\label{fig-alg-stream}
\vspace{-2mm}
\end{figure}

Our idea is to treat the node set of $G$ as a stream, 
and incrementalize the update of 
``lower-tier'' explanation graph $G^l_s$ 
and {\em accordingly} the ``affected'' higher-tier patterns $\P$, 
to reduce unnecessary verification. 
To this end, it uses the following  
procedures: (1) \kw{IncEVerify} and \kw{IncPMatch}: 
Upon the arrival of a node $v$, 
\kw{IncEVerify} only updates 
the feature influence $I_1(.,v)$, 
diversity $D(v)$, and updates $I(V_S\cup\{v\})$ 
and $D(V_S\cup\{v\})$ incrementally; 
\kw{IncPMatch} invokes fast 
incremental and streaming subgraph
matching algorithms, \eg~\cite{fan2013incremental, kim2018turboflux} 
to check graph views, explanation views, 
and proper coverage.   
(2) \kw{IncPGen}: unlike \pgen, it 
takes as input a small subgraph 
induced by the $r$-hop neighbors of $v$ (where $r$ is specified in $\C$ for 
neighborhood influence), and 
only generates new patterns $\Delta \P$
not in $\P^l$, to be verified by \kw{IncPMatch}; 
(3) \kw{IncUpdateP} and \kw{IncUpdateVS} maintain 
$\P^l$ and $G^l_s$ respectively, 
with a space-efficient ``swapping''  
strategy (to be discussed) to dynamically decide 
whether to replace patterns.

\stitle{Algorithm.} 
The algorithm, denoted as  \gvstream, is outlined in Algorithm \ref{alg:streaming_selection}. 
Upon the arrival of a node $v$, 
it invokes \kw{IncEVerify} to 
maintain the Jacobian matrix,
updates the marginal gain, and enlarges 
the candidate set $V_u$. 
It then tests the extendability of 
$V_S$, and invokes \kw{IncUpdateVS} and 
\kw{IncUpdateP} to update  
$\P^l$ and $\G^l_s$ respectively. 
The ``post processing'' is similar 
as its counterpart in \gvapprox (line~10) 
ensures the 
lower bound.

\eetitle{Local Incremental Update.} 
\kw{IncUpdateVS} maintains 
$V_S$ as a node cache of size 
up to $\C.u_l$. 
For a node $v$ that 
passes extendable test (line~6), 
it consults a greedy swapping strategy to decide whether to replace a node $v'\in V_S$ with $v$ or reject $v$, 
and put the node $v'$ into $V_u$. Specifically, 
it performs a case analysis: 
(a) if $|V_S|\textless \C.u_l$, it simply 
adds $v$ to $V_S$;  
(b) otherwise, if $\P^l$ already 
covers $v$, or $v$ alone 
does not contribute new 
patterns to $\P^l$ ($\Delta\P = \emptyset$, 
as determined by \kw{IncPGen}), 
it skips processing $v$, as this 
does not hurt the quality of the current 
explanation view; 
(b) otherwise, it 
chooses the node $v'\in V_S$ 
whose removal has the smallest ``loss'' of 
explainability score, 
and replaces $v'$ with $v$ only 
when such a replacement 
ensures a gain that is at least twice as much as 
the loss. In other words, 
the replacement does not hurt  
the original approximation ratio. 

Upon the formation of new 
explanation subgraphs, \kw{IncUpdateP} 
performs a similar case analysis, 
yet on patterns $\P^l$, and conducts a  
swapping strategy to 
ensure node coverage and 
small edge misses. 
We present the details of 
~\kw{IncUpdateVS} and 
\kw{IncUpdateP} in the appendix.  

\begin{example}
Figure \ref{fig-alg-stream} illustrates 
the maintenance of an explanation view with 
four explanation subgraphs ($\{G_{s7},G_{s8},G_{s9},G_{s10}\}$ that are properly covered by four patterns $\{P_{41},P_{42},P_{43},P_{44}\}$). 
Upon the processing of a new node, a newly induced explanation 
subgraph $G_{s11}$ is to be processed. As existing patterns $\{P_{41},P_{42},P_{43},P_{44}\}$ already cover a fraction of $G_{s11}$, \gvstream masks the nodes that are covered and proceeds to generate a new pattern $P_{45}$ 
from $G_{s11}$ to cover its remaining fraction. 
The explanation view is eventually enriched with $G_{s11}$ and a new pattern $P_{45}$.
\end{example}

\vspace{-2ex}
\stitle{Analysis}. 
The approximation guarantee of \gvstream comes from the $\frac{1}{4}$-approximation ensured by the streaming submodular maximization~\cite{halabi2020fairness}, as well as the online optimization of full coverage during the explanation generation. Specifically, its greedy local replacement strategy ensures an invariant that the selected nodes do not impact the $\frac{1}{4}$ approximation ratio. Online pattern generation does not affect the full coverage property, thus assuring quality. 
\gvstream offers the advantage of not requiring a comparison of information from all nodes each time, allowing anytime access of explanation views. As the processing is performed ``one node at a time'', the pattern generation is expedited, further enhancing its speed. 

\gvstream does not require any prior node order. {\em (i)} It ensures “anytime” quality guarantees regardless of node orders (Theorem \ref{thm-online}). {\em (ii)} Prioritizing some nodes may allow the early discovery of certain frequent patterns. \kw{IncUpdateP} maintains $\P^l$ with a space-efficient ``swapping'' strategy to dynamically decide whether to replace patterns and nodes. Thus, the higher-tier patterns may vary slightly under different node orders, though a significant majority of the important patterns captured will be similar. {\em (iii)} Different node orders in \gvstream does not affect the worst case time cost.

\eetitle{Parallel Implementation}. 
To generate $\G_\V$ over $\G$ with 
multiple labels, one can readily apply   
a parallel scheme with  
$|\G|$ processes, each processes 
a node stream by invoking \gvstream. 
We present a detailed analysis in the appendix.

\vspace{-0.8mm}
\section{Experimental Study}
\label{sec:exp}
We conduct an empirical evaluation of our solutions and existing approaches using both real-world and synthetic graph databases (Table \ref{tab:prop}). All methods are implemented in Python. The experiments are executed on a Ubuntu machine with one NVIDIA GeForce RTX 3090 GPU and 128G RAM on Intel(R) Xeon(R) CPU E5-2650 v4 @ 2.20GHz CPU. We employ multi-processing to demonstrate the parallelism of our algorithms. Our code and datasets are at \cite{code}.

\begin{table}[htb!]
    \centering
    \vspace{-1mm}
    \small
    \caption{Dataset statistics. NF inidicates node features.}
    \vspace{-3mm}
    
    \begin{tabular}{c || c c c c c}
            \multirow{2}{*}{\textbf{Dataset}}  & {\bf Avg \# Edges} & {\bf Avg \# Nodes} & {\bf \# NF} & {\bf \# Graphs} & {\bf \# Classes} \\
            &
            {\textbf{per graph}}  & {\bf per graph} & {\bf per node} &  &  \\
            \midrule

		  {\bf MUTANGENICITY} & 31 & 30 & 14 & 4337 & 2 \\ 

		  {\bf REEDIT-BINARY} & 996 & 430 & - & 2000 & 2 \\ 
            {\bf ENZYMES} & 62 & 33 & 3 & 600 & 6 \\ 
		  {\bf MALNET-TINY} & 2860 & 1522 & - & 5000 & 5 \\ 
            {\bf PCQM4Mv2} & 31 & 15 & 9 & 3\,746\,619 & 3 
          \\
            {\bf PRODUCTS} & 5\, 728\,239 & 1\,184\,330 & 100 & 1 & 47 
          \\
            {\bf SYNTHETIC} & 1\,999\,975 & 400\,275 & - & 100 & 2 
    \end{tabular}
    \label{tab:prop}
    \vspace{-3mm}
\end{table}

\vspace{-1.5mm}
\subsection{Experimental Setup}

\vspace{-2mm}
\stitle{Datasets.} {\bf (1) MUTAGENICITY ({\sf MUT})  }\cite{kazius2005derivation} is a molecular dataset for binary classification task. Each graph represents a chemical compound, where nodes are atoms and undirected edges denote bonds. The one-hot node feature indicates the atom type, e.g., carbon, oxygen.
{\bf (2) REDDIT-BINARY ({\sf RED})} \cite{yanardag2015deep} is a social network dataset comprising 2000 online discussion threads on Reddit. Nodes are users participating in a certain thread, while an edge denotes that a user responded to another. These graphs are labeled based on two types of user interactions, {\em question-answer} and {\em online-discussion}, in the threads.
{\bf (3) ENZYMES {\sf (ENZ)}}  \cite{borgwardt2005protein} is a protein dataset, containing hundreds of undirected protein-protein interaction structures for up to six types of enzymes. One-hot node features indicate the type of protein. 
{\bf (4) MALNET-TINY ({\sf MAL})} ~\cite{freitas2020large} is an ontology of malicious software function call graphs (FCGs). Each FCG captures calling relationships between functions within a program, with nodes representing functions and directed edges indicating inter-procedural calls. The individual graph size in this dataset is considerably larger, posing additional challenges for identifying concise explanation substructures.
{\bf (5) PCQM4Mv2 {\sf (PCQ)}} ~\cite{huogb} is a quantum chemistry dataset originally curated under the PubChemQC project. It provides molecules as the SMILES strings, from which 2D molecule undirected graphs (nodes are atoms and edges are chemical bonds) are constructed, where each node is associated with a 9-dimensional feature fingerprint. 
{\bf (6) PRODUCTS ({\sf PRO})} \cite{HuFZDRLCL20} represents an Amazon product co-purchasing network and consists of an undirected, unweighted graph containing 2,449,029 nodes and 61,859,140 edges. The task is to predict the category of a product, where the 47 top-level categories are used as target labels. Originally it was designed for node classification and we transform this dataset for a graph classification task by sampling 400 subgraphs from the original graph.
{\bf (7) SYNTHETIC ({\sf SYN})} is a synthetically generated graph dataset through the PyTorch Geometric library. This dataset leverages the {\sf BA}-graph ({\sf Barabasi-Albert} graph) as its base graph and incorporates {\sf HouseMotif} and {\sf CycleMotif} as motif generators, each assigned to separate two classes \cite{ying2019gnnexplainer}. One single graph of this dataset contains approximately 0.4 million nodes and 2 million edges.

\vspace{-0.8mm}
\stitle{Classifier.} 
In line with recent works \cite{yuan2021explainability,ying2019gnnexplainer,zhang2022gstarx,huang2023global}, we employ a classic message-passing \gnn, namely a graph convolutional network (GCN) with three convolution layers, each having an embedding dimension of 128. To facilitate classification, the GCN model is enhanced with a max pooling layer and a fully connected layer. For datasets without node features, we assign each node a default feature. During training, we utilize the Adam optimizer \cite{kingma2014adam} with a learning rate of 0.001 for 2000 epochs. The datasets are split into 80\% for training, 10\% for validation, and 10\% for testing. The explanations are generated based on the classification results of the test set. Recall that our proposed solutions are model-agnostic, making them adaptable to any \gnn employing message-passing.

\vspace{-0.8mm}
\stitle{Competitors.} To our best knowledge, \gvex is the first configurable label-level explainer. To demonstrate its effectiveness, we compare it with 4 state-of-the-art \gnn explainers, making minor adjustments as necessary to ensure fair comparison. We denote our two-step method as {\sf ApproxGVEX (AG)} (\S\ref{sec-approx}) and our steaming method as {\sf SteamGVEX (SG)} (\S\ref{sec-stream}). \textbf{(1) GNNExplainer {\sf (GE)}} ~\cite{ying2019gnnexplainer} learns soft masks based on mutual information to select critical edges and node features that influence instance-level classification results. 
\textbf{(2) SubgraphX {\sf (SX)}} ~\cite{yuan2021explainability}  employs the Monte Carlo tree search to efficiently explore different subgraphs via node pruning and select the most important subgraph as the explanation for instance-level graph classification. 
\textbf{(3) GStarX {\sf (GX)}} ~\cite{zhang2022gstarx}  designs node importance scoring functions using a new structure-aware value from cooperative game theory. It identifies critical nodes and generates an induced subgraph as the explanation for each input graph.
\textbf{(4) GCFExplainer {\sf (GCF)}} ~\cite{huang2023global} explores the global explainability of \gnns through counterfactual reasoning. It identifies a set of counterfactual graphs that explain all input graphs of a specific label.

We do not compare against {\sf XGNN} \cite{yuan2020xgnn} and {\sf PGExplainer} \cite{luo2020parameterized} since (1) unlike ours and above competitors, {\sf XGNN} is a model-level explainer, it does not rely on input graphs to generate explanations. As a result, calculating fidelity (see below) becomes difficult. 
(2) {\sf PGExplainer} is similar to {\sf GNNExplainer}, it focuses on edge-level explanation rather than subgraph-level and is not a black box. Therefore, we opted for the more representative method, {\sf GNNExplainer}.

\stitle{Evaluation metrics}. We evaluate the quality of explanations considering explanation faithfulness and conciseness.

\eetitle{Explanation faithfulness.} {\sf Fidelity+} and {\sf Fidelity-} ~\cite{yuan2022explainability} are two widely-used metrics for assessing if explanations are faithful to the model, that is, capable of identifying input features important for the model. Fidelity+ quantifies the deviations caused by a targeted intervention, i.e., removing the explanation substructure from the input graph.
\begin{small}
\begin{equation}
    Fidelity+ = \frac{1}{|\G|} \mathop{\sum}_{G\in \G}(Pr(\M(G)=l_G)-Pr(\M(G')=l_G))
\end{equation}
\end{small}

\noindent where $l_G$ is the original prediction for the graph $G$. $G'$ represents the updated graph obtained by masking the explanation substructure from the original graph $G$.
Fidelity+ metric measures the difference in probabilities between the new predictions and the original ones. A higher Fidelity+ score indicates better distinction.

In contrast, Fidelity- metric measures how close the prediction results of the explanation substructures are to the original inputs.
\begin{small}
\begin{equation}
    Fidelity- = \frac{1}{|\G|} \mathop{\sum}_{G\in \G}(Pr(\M(G)=l_G)-Pr(\M(G_s)=l_G))
\end{equation}
\end{small}
A desirable Fidelity- score should be close to or even smaller than zero, indicating perfect-matched or even stronger predictions.

We evaluate the explainability of the subgraphs in our explanation views. As clarified earlier in Section~\ref{sec:gnn_views}, the ``lower-tier'' subgraphs are responsible for explaining \gnns with consistent (Fidelity-) and counterfactual (Fidelity+) properties. On the other hand, the ``higher-tier'' patterns are provided to facilitate better query-ability, as assessed through the following metrics.

\eetitle{Conciseness.} 
To assess the conciseness of explanation subgraphs produced by ours and various competitors, we employ the well-known {\sf sparsity} metric \cite{yuan2022explainability}, 
computed as:
\begin{small}
\begin{equation}
 Sparsity = \frac{1}{|\G|} \mathop{\sum}_{G\in \G}(1 - \frac{|V_s|+|E_s|}{|V|+|E|})  
\end{equation}
\end{small}
where the nodes and edges in the input graph $G$ and its explanation subgraph $G_s$ are denoted by $(V,E)$ and $(V_s,E_s)$, respectively.
Higher Sparsity values indicate more concise explanations.

Finally, we assess the {\sf compression} due to ``higher level'' explanation patterns, which act as summaries of the ``lower level'' subgraphs. This metric is applicable only for our two-tier explanation views, where the nodes and edges of explanation subgraphs and patterns are denoted as $(\mathbf{V}_{S},\mathbf{E}_{S})$ and $(\mathbf{V}_{\P},\mathbf{E}_{\P})$, respectively.
\begin{small}
\begin{equation}
    Compression = 1-\frac{|\mathbf{V}_{\P}| +|\mathbf{E}_{\P}|}{|\mathbf{V}_{S}| + |\mathbf{E}_{S}|}
\end{equation}
\end{small}

\subsection{Experimental Results} 
\vspace{-1mm}

\stitle{Exp-1: Effectiveness}. Below we report the effectiveness of \gvex. 
\label{sec:effectiveness}
\vspace{-1mm}

\eetitle{Explanation faithfulness.}  
To validate the consistency and counterfactual nature of our explanation subgraphs, we generate explanations for one label of user's interest, and vary the configuration constraint $u_l$ to control the maximum number of nodes in explanation subgraphs. 
For competitors, as they do not have configurable options, we consider their overall qualities for this specific label. 
If a competitor is absent in the evaluation of a dataset, it indicates that the method took a longer time, i.e., $>$ 24 hours, on that dataset.

\begin{figure}[t!]
	\vspace{-4mm}
	\centering

	\subfigure[{\sf RED}]
	{\includegraphics[width=0.23\linewidth]{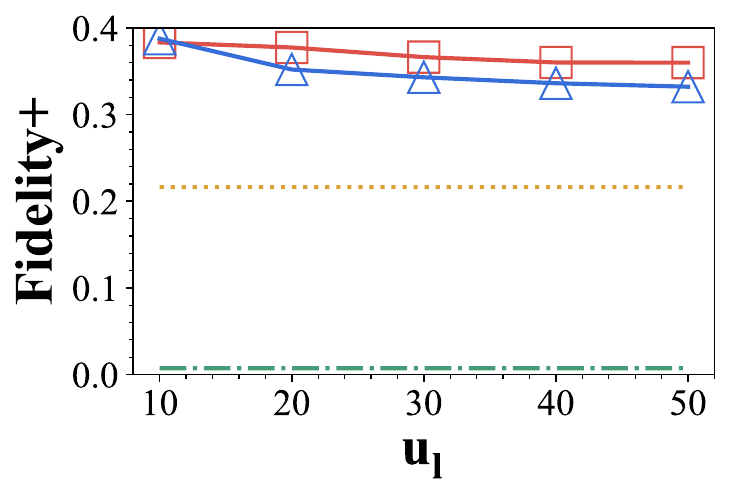}
	    \label{fig:re-fide+}}
    \subfigure[{\sf ENZ}]
	{\includegraphics[width=0.23\linewidth]{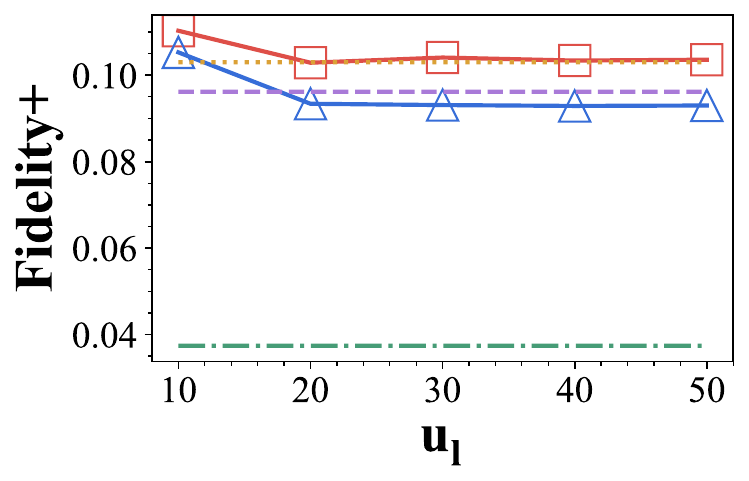}
		\label{fig:en-fide+}}	
	\subfigure[{\sf MUT}]
	{\includegraphics[width=0.23\linewidth]{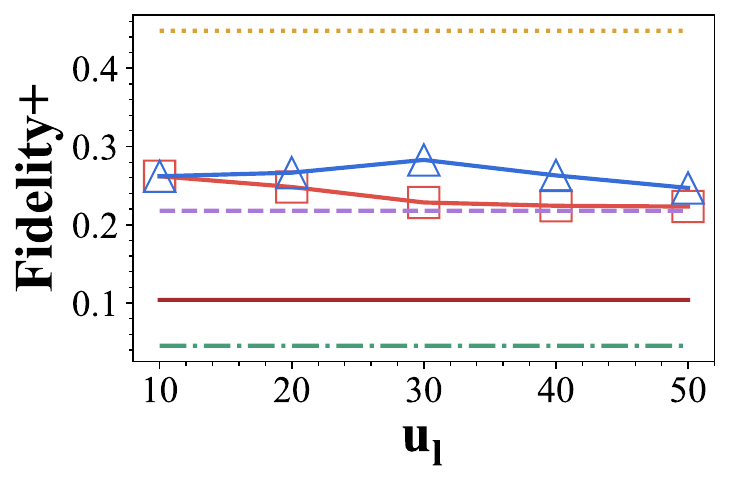}
		\label{fig:mutag_fide+}}
    \subfigure[{\sf MAL}]
	{\includegraphics[width=0.23\linewidth]{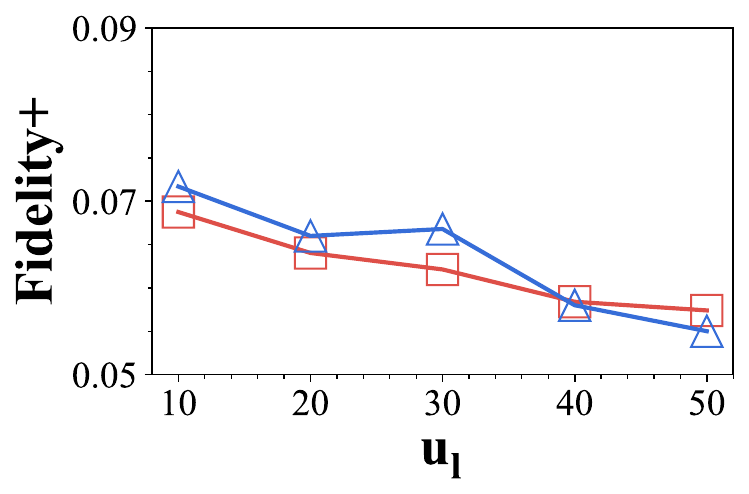}
		\label{fig:mal-fide+}}
	\vspace{-4.5mm}
	\caption{The Fidelity+ comparison across various GNN explainers under different configuration constraints}
	\label{fig:fide-}
	\vspace{-3.5mm}
\end{figure}

\begin{figure}[t!]
	\vspace{-2mm}
	\centering

	\subfigure[{\sf RED}]
	{\includegraphics[width=0.23\linewidth]{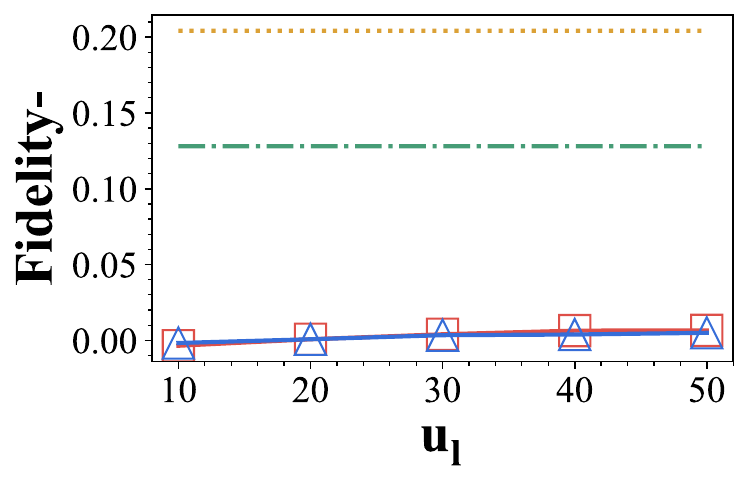}
	    \label{fig:re-fide-}}
    \subfigure[{\sf ENZ}]
	{\includegraphics[width=0.23 \linewidth]{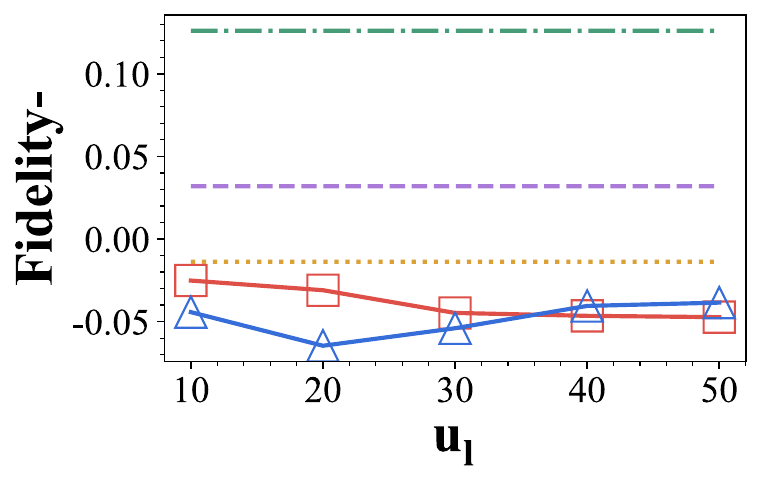}
		\label{fig:en-fide-}}	
  	\subfigure[{\sf MUT}]
	{\includegraphics[width=0.23\linewidth]{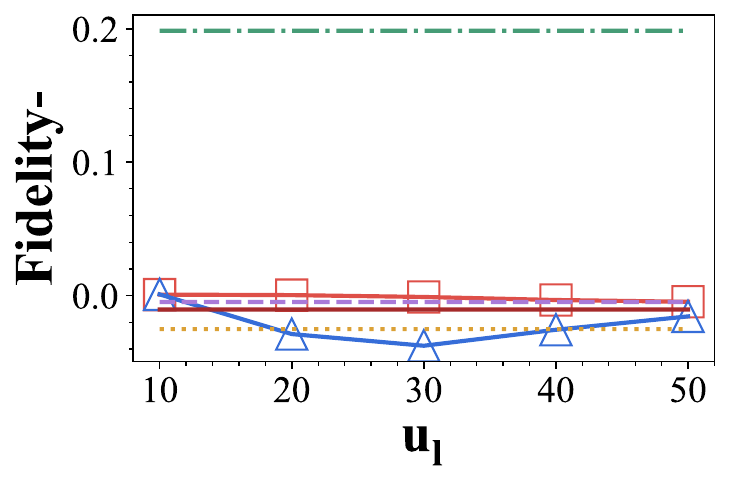}
		\label{fig:mutag_fide-}}
    \subfigure[{\sf MAL}]
	{\includegraphics[width=0.23 \linewidth]{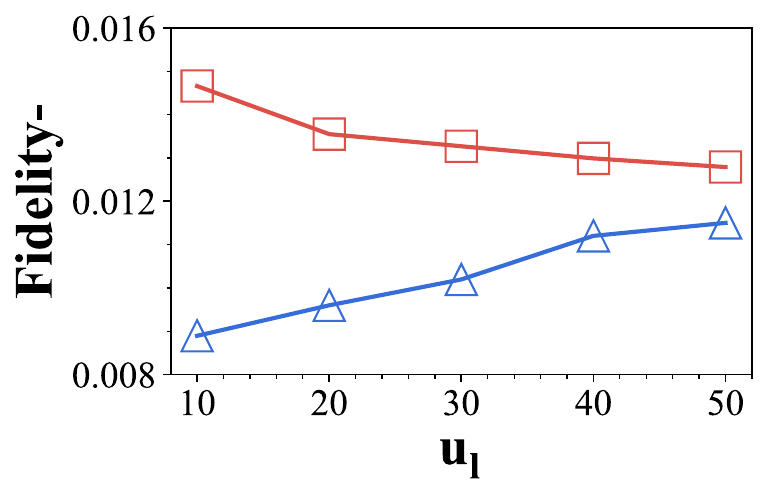}
		\label{fig:mal-fide-}}
	\vspace{-4.5mm}
	\caption{The Fidelity- comparison across various GNN explainers under different configuration constraints} 
	\label{fig:fide+}
	\vspace{-3.5mm}
\end{figure}

Figure \ref{fig:fide-} and Figure \ref{fig:fide+} showcase the fidelity metrics with varying $u_l$. Notably, our proposed \gvapprox and \gvstream methods consistently outperform all other competitors. They achieve higher Fidelity+ scores (consistent) on all datasets (except for the {\sf MUT} dataset) and lower Fidelity- scores (counterfactual) on all datasets. Unlike {\sf  GNNExplainer} and {\sf SubgraphX}, our objective in capturing explainability (Eq.~\ref{eq:goodness}) focuses on feature influence and diversity, rather than explicitly optimizing for differences with respect to original prediction results.  
This shows that our extracted message-passing substructures indeed carry critical information that faithfully corresponds to the classification results.

\gvapprox and \gvstream have minor quality gaps up to 0.023 (Fidelity). \gvstream is more fluctuating than \gvapprox. For instance, in {\sf MAL}, the effectiveness of \gvstream diminishes more rapidly, falling behind \gvapprox when the explanation sizes becomes larger.
In contrast, \gvapprox maintains a more consistent and uniform trend and performs better due to tighter approximation.

The parameter $u_l$ enforces an upper bound on the size of explanations. Thus, a larger $u_l$ leads to more comprehensive explanations for a class at the expense of higher time cost.

Furthermore, on {\sf MUT} dataset, we vary the parameters to observe how the fidelity values respond to various combinations of $(\theta, r)$. We also adjust the values of $\gamma$ for the fixed $(\theta, r)$ combinations.
$\theta$ is used to control the boundary of feature influence, $r$ controls the  neighborhood diversity, $\gamma$ is a trade-off between the two. 
The parameter setting is optimized by grid search. 
For {\sf MUT} dataset, we set $(\theta,r)$ to $(0.08, 0.25)$ and $\gamma$ to $0.5$.
This serves the dual purpose of enabling the algorithm to identify influential nodes possessing diversity while striking a suitable balance between them.

\begin{figure}[t!]
  \vspace{-3mm}
  \centering
  \includegraphics[width=0.7\linewidth]{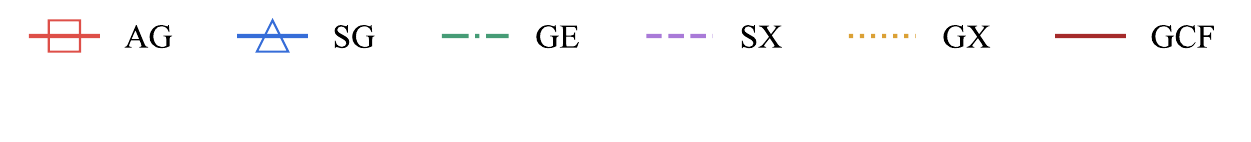}
  \vspace{-7mm}
  \label{fig:legends}
    \vspace{-1mm}
\end{figure}

\begin{figure}[t!]
	\vspace{-2mm}
	\centering
	\subfigure[Fidelity+ via $(\theta,r)$]
	{\includegraphics[width=0.23\linewidth]{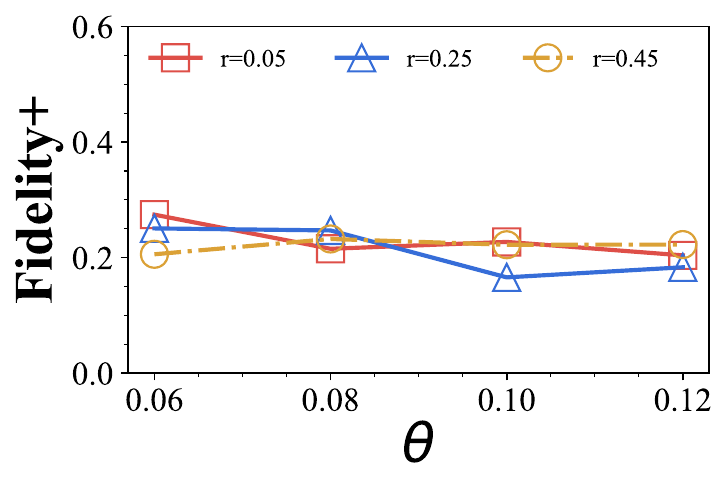}
		\label{fig:config_fide+}}
	\subfigure[Fidelity- via $(\theta,r)$]
	{\includegraphics[width=0.23\linewidth]{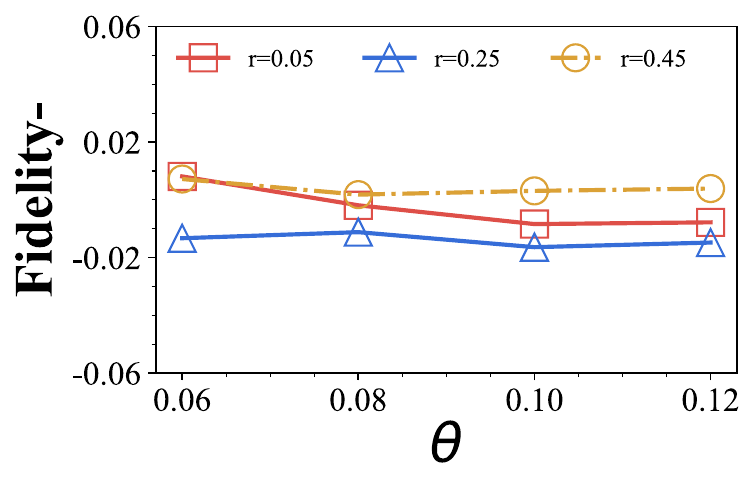}
	    \label{fig:config-fide-}}
    \subfigure[Fidelity+ via $\gamma$]
	{\includegraphics[width=0.23\linewidth]{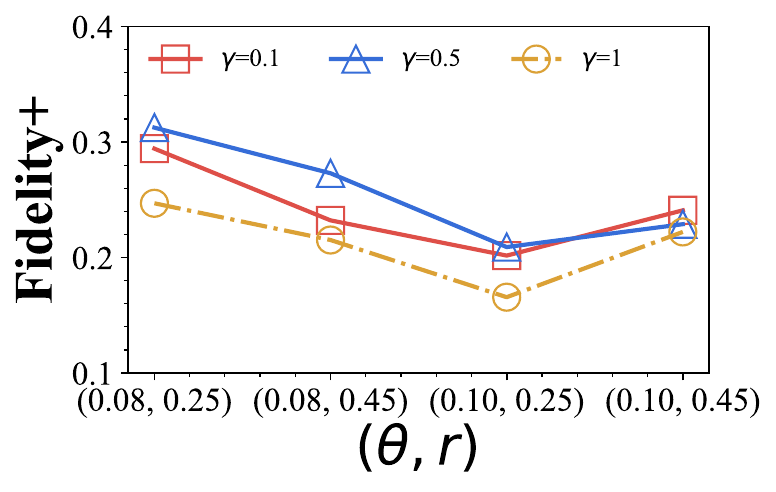}
		\label{fig:gamma-fide+}}	
    \subfigure[Fidelity- via $\gamma$]
	{\includegraphics[width=0.23\linewidth]{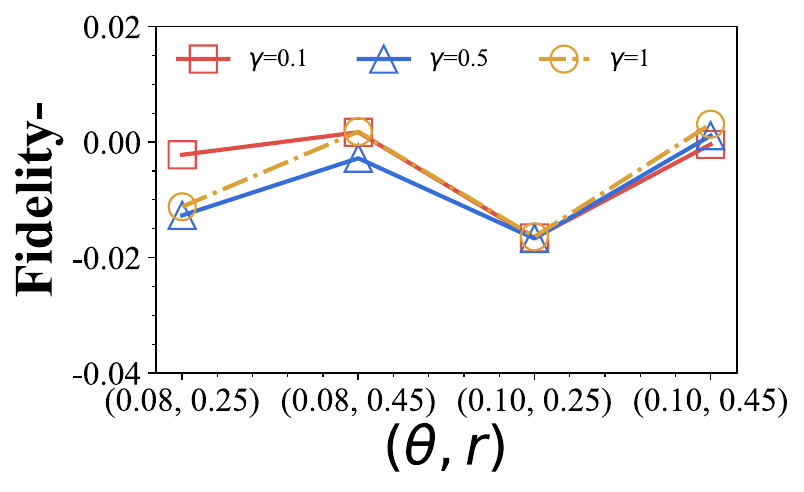}
		\label{fig:gamma-fide-}}
	\vspace{-4.5mm}
	\caption{Configuration parameters anayses}
	\label{fig:configuration}
	\vspace{-3.5mm}
\end{figure}

\begin{figure}[tb!]
	\vspace{-2mm}
	\centering
        \subfigure[Sparsity]
	{\includegraphics[width=0.23 \linewidth]{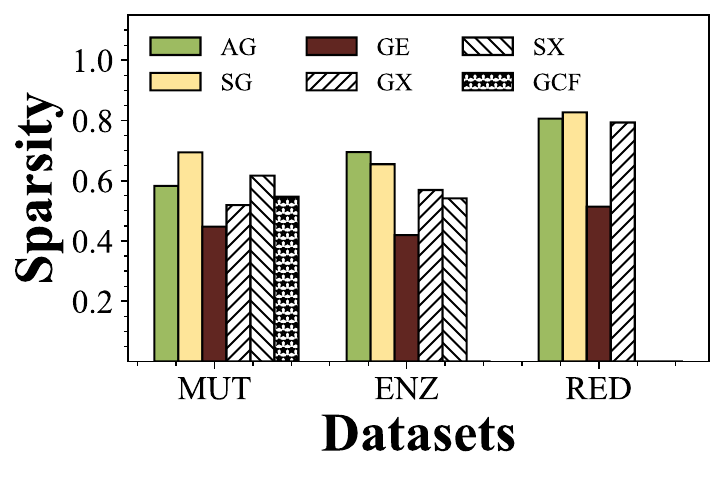}
	    \label{fig:sparsity}}
	\subfigure[Compression]
	{\includegraphics[width=0.23 \linewidth]{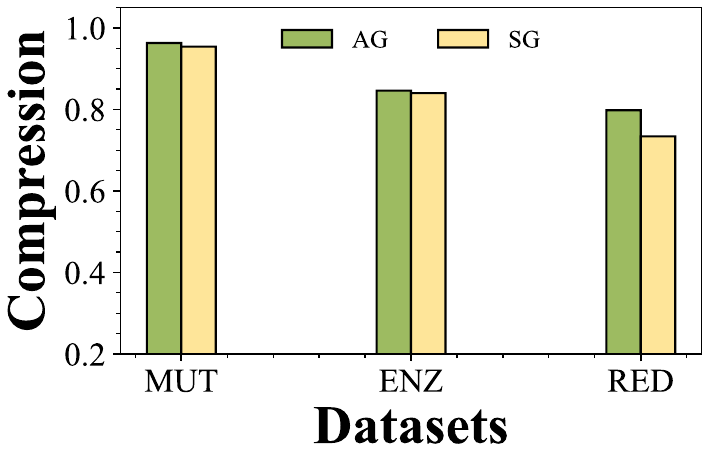}
    \vspace{-3ex}
		\label{fig:compression-ratio}}
    \subfigure[Edge Loss ({\sf MUT})]
  	{\includegraphics[width=0.23 \linewidth]{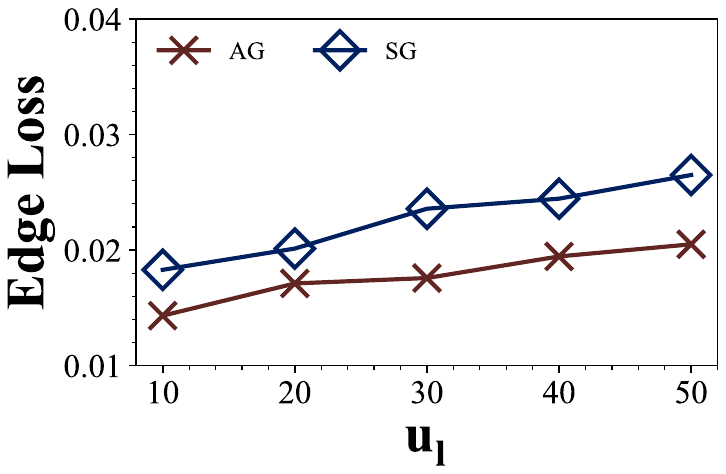}
		\label{fig:edge-mut}}
    \subfigure[Edge Loss ({\sf RED})]
  	{\includegraphics[width=0.23 \linewidth]{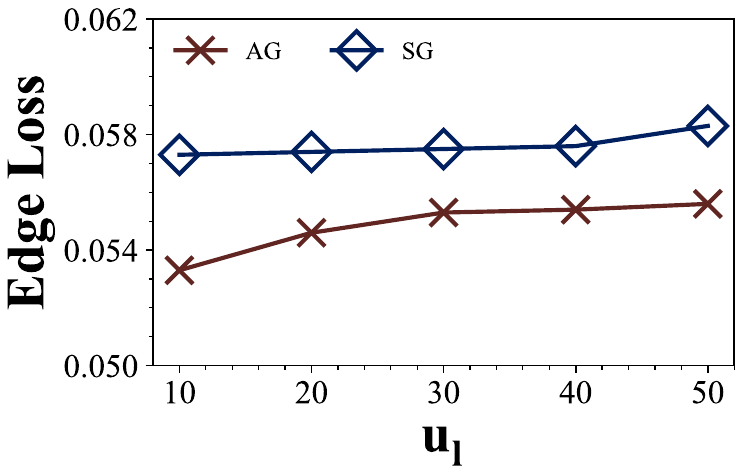}
		\label{fig:edge-red}}
	\vspace{-4.5mm}
	\caption{Conciseness analyses}
	\label{fig:two-coverage}
        \vspace{-3mm}
\end{figure}

\eetitle{Conciseness.} Figure~\ref{fig:sparsity} depicts the results of sparsity. It is evident that both \gvapprox and \gvstream consistently generate more compact explanation subgraphs across all datasets. The performance gap can be as high as 0.2 compared to {\sf GNNExplainer}, which fails to effectively prune unessential topological structures. Overall, \gvapprox and \gvstream  significantly reduce the total number of nodes and edges by 60\% to 80\%, and retain important information to be explored by human experts. \gvapprox and \gvstream differ only slightly on all datasets because our configuration parameters bound the number of nodes in explanations, which in turn produces slight differences in the number of edges in explanations generated by the two algorithms.

Figure~\ref{fig:compression-ratio} demonstrates an excellent reduction in the number of nodes and edges achieved by our ``higher-tier'' patterns relative to ``lower-tier'' subgraphs. It reveals that more than 95\% of nodes can be further compressed. Recall that our algorithms ensure
full coverage of the nodes in the
explanation subgraphs by patterns set via node-induced subgraph isomorphism. This observation highlights that the explanation subgraphs can be effectively represented by several significantly smaller substructures. Furthermore, our case study shows that the patterns exhibit significant variation when the labels of interest change. These indicate that \gvex can identify both compact and highly informative patterns, enabling domain experts to explore the critical information from the graphs.

Figure~\ref{fig:edge-mut},~\ref{fig:edge-red} show the the impact of $u_l$ on edge loss. Edge loss is the percentage of edges that our high-tier patterns fail to cover in the low-tier explanation subgraphs while we satisfy the node coverage constraints in $\C$ (see Lemma~\ref{lm-coverage}). We vary the configuration constraint $u_l$ to control the maximum number of nodes in explanation subgraphs. It depicts that the percentage of edges that the algorithm failed to cover increases when $u_l$ increases. Specifically, in {\sf MUT} dataset, as $u_l$ varies, the percentage of edges remaining uncovered manifests as $\{1.43\%, 1.71\%, 1.75\%, 1.95\%, 2.10\%\}$.

\begin{figure}[tb!]
	\vspace{-3mm}
	\centering
  \subfigure[Runtime ({\sf MUT})]
	{\includegraphics[width=0.32 \linewidth]{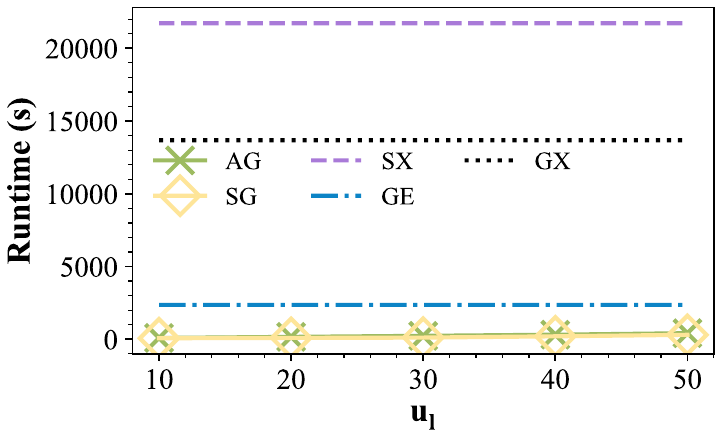}
		\label{fig:budget-mut-eff}}    
  \subfigure[Runtime ({\sf ENZ})]
	{\includegraphics[width=0.32 \linewidth]{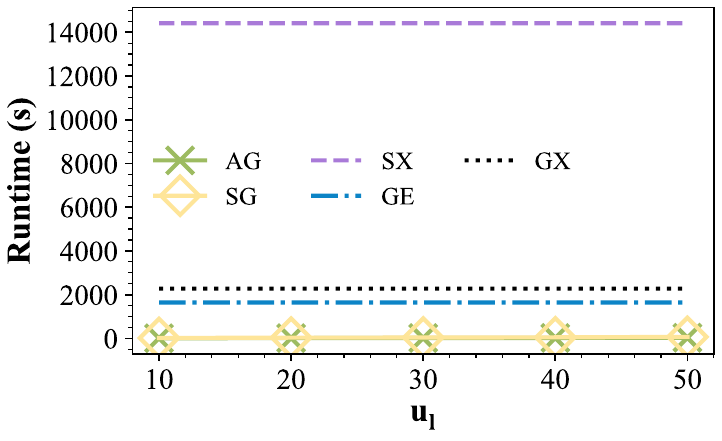}
		\label{fig:budget-enz-eff}}
    \subfigure[Runtime]
	{\includegraphics[width=0.32 \linewidth]{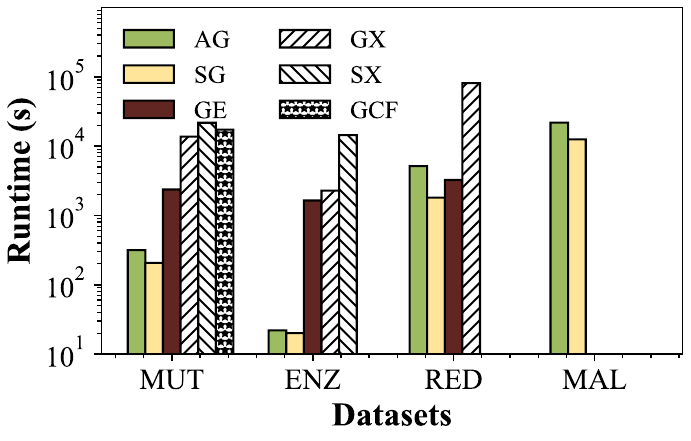}
		\label{fig:overall-eff}}	
        \subfigure[Scalability (PCQ)]
	{\includegraphics[width=0.32 \linewidth]{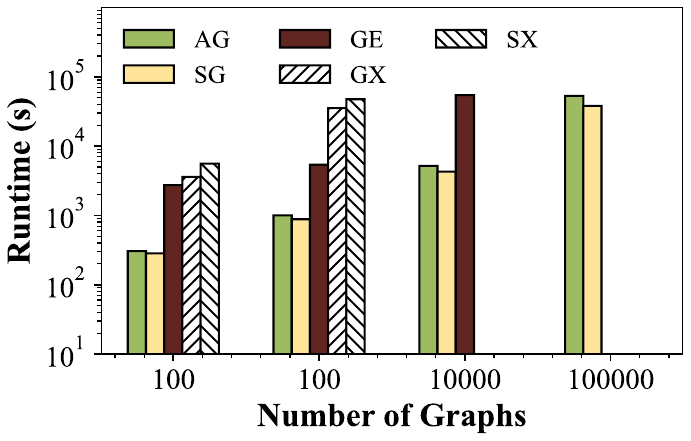}
    \label{fig:scala-eff}}
         \subfigure[Parallelization]
	{\includegraphics[width=0.32 \linewidth]{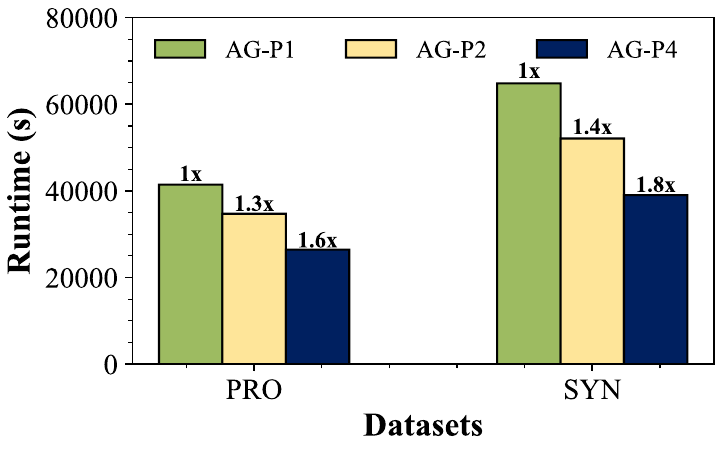}
		\label{fig:parallel-eff}}
    \subfigure[Anytime Efficiency (PCQ)]
	{\includegraphics[width=0.32 \linewidth]{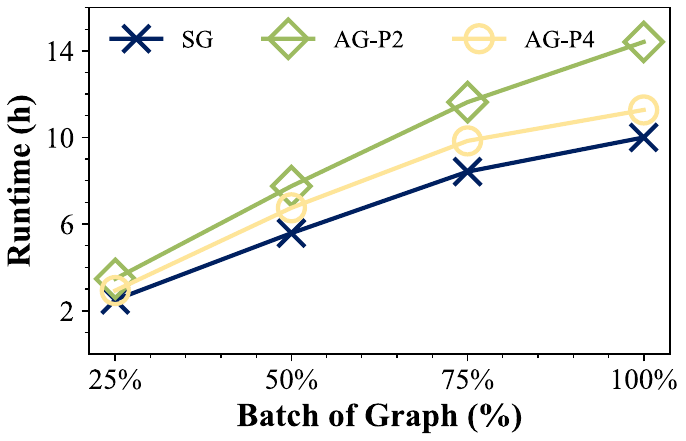}
		\label{fig:delay-time}}
        \vspace{-4mm}
	\caption{Efficiency, scalability, and parallelization analyses}
	\label{fig:effi-exp}
	\vspace{-5mm}
\end{figure}

\stitle{Exp-2: Efficiency and Scalability}. 
\label{sec:efficiency}
We next demonstrate that our methods consistently generate graph explanations in a more efficient manner, even when dealing with graph databases that have relatively larger individual graphs (e.g., {\sf PRO}, {\sf SYN}, {\sf MAL}) or a large number of graph instances (e.g., {\sf PCQ}). Figures ~\ref{fig:budget-mut-eff}-\ref{fig:budget-enz-eff} present the running times of our \gvapprox and \gvstream methods, showcasing their significantly faster performance compared to various competitors by 1-2 orders of magnitude. Both \gvapprox and \gvstream complete their execution within hundreds of seconds on {\sf MUT} and {\sf ENZ}, providing substantial improvements in efficiency.

Figure ~\ref{fig:overall-eff} provides a more comprehensive overview of the running times of all explainers across various datasets. Notice that all competitors are absent in {\sf MAL} dataset, which contains relatively larger individual graphs. Additionally, when considering more input graphs on the {\sf PCQ} dataset, all competitors require $>$ 24 hours with 100k graphs as shown in Figure ~\ref{fig:scala-eff}. In contrast, \gvex successfully completes the task in approximately 8 hours with 100k graphs. These demonstrate the superiority of \gvex solution's in scalability in terms of relatively larger as well as more graphs.

Figure ~\ref{fig:parallel-eff} shows that our running time reduces by nearly 2$\times$ with 
parallel processing. For {\sf PRO} dataset, we observe that a node's classification is influenced by message-passing among its neighboring nodes. So we adopt a strategy where we select a specific number of nodes and consider their neighboring nodes to construct subgraphs. The label assigned to a node becomes the label for the entire subgraph. We sample approximately 400 subgraphs, each containing roughly 3000 nodes, resulting in a subgraph classification task involving approximately 1 million nodes and 6 million edges. It takes \gvex about 7 hours to complete this task. For {\sf SYN} dataset, we use sparse
matrix multiplication and random walk technique~\cite{cohen1999approximating,avrachenkov2007monte} to optimize the
computation on large graphs, and parallelize 
on multi-processes. With 4 processes, \gvex successfully completes the task in approximately 10 hours. These results demonstrate the efficiency and scalability of the \gvex algorithm when confronted with large, connected graph datasets.

Finally, our streaming method, \gvstream, exhibits linear growth in running time with batch size, measued by the percentage of test graphs, making it highly scalable (Figure ~\ref{fig:delay-time}). It also remains more efficient than the 4-processor parallel version of \gvapprox, emphasizing its suitability for handling large-scale graph datasets.

\begin{figure}[t!]
	\centering
	\includegraphics[width=1 \linewidth]{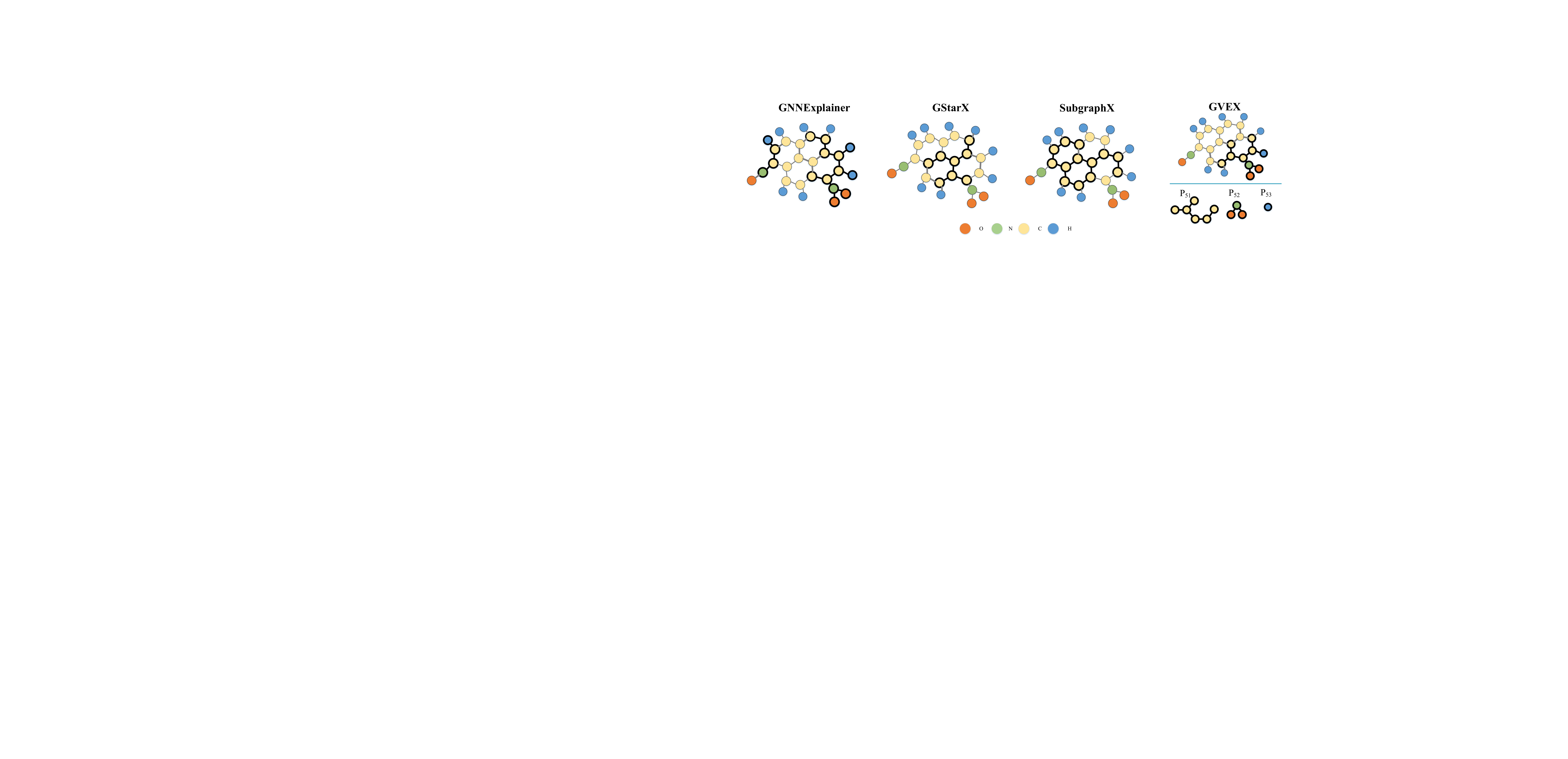}
        \vspace{-6mm}	
	\caption{Case study 1 on \gnn-based drug design}
	\label{fig:case-study}
        \vspace{-3mm}
\end{figure}

\begin{figure}[t!]
	\centering
	\includegraphics[width=0.91 \linewidth]{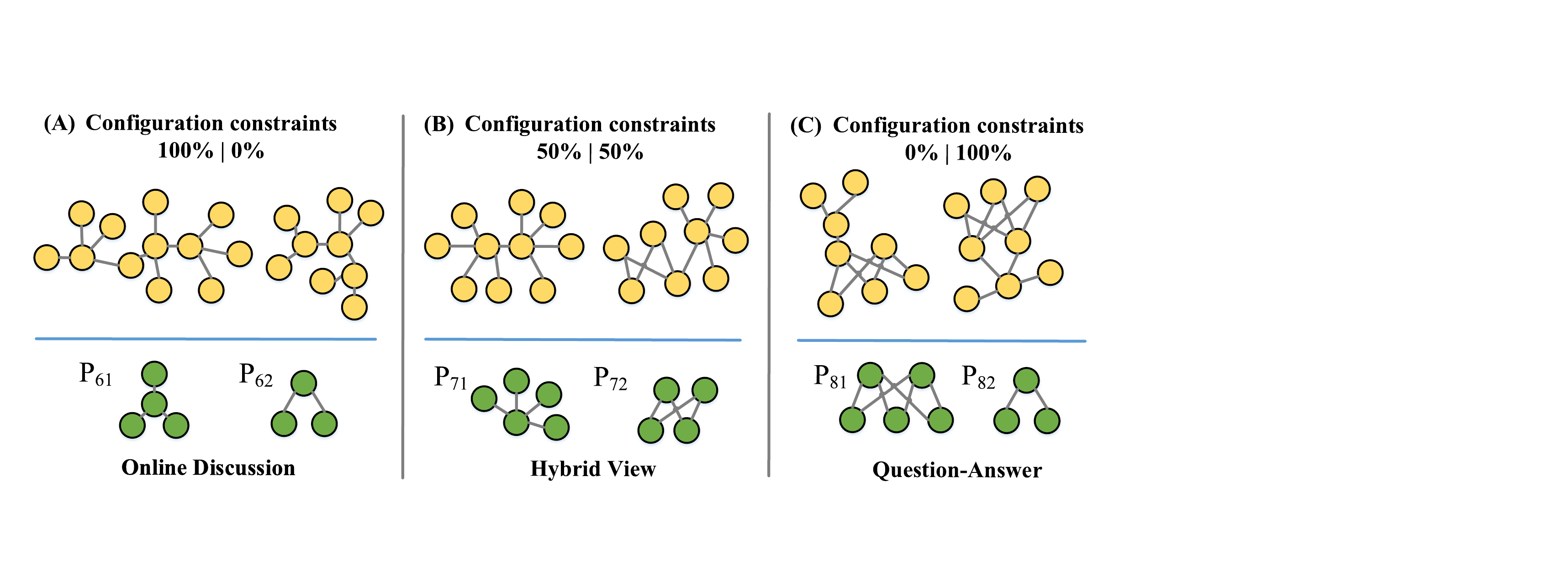}
        \vspace{-4mm}	
	\caption{Case study 2 on \gnn-based social analysis}
	\label{fig:case-study-2}
       \vspace{-2mm}
\end{figure}

\stitle{Exp-3: Case Studies}. 
\label{sec:casestudy}
In our first case study, we compare the explanation subgraphs identified for one mutagen by different explainers, highlighting them with thicker lines on the input graph (Figure~\ref{fig:case-study}). It is evident that \gvex produces smaller subgraphs compared to {\sf GNNExplainer} and {\sf SubgraphX}. Furthermore, our explanation view breaks down such subgraphs into smaller components that may appear multiple times, facilitating easier access and exploration.
\gvex successfully identifies the real toxicophore, $NO_2$, allowing for correct and efficient query answering in downstream analytical tasks such as {\em ``which toxicophore occurs in mutagens?''}. Among the competitors, only {\sf GNNExplainer} includes $NO_2$ in its output, albeit with an explanation subgraph consisting of 14 atoms.

Figure \ref{fig:case-study-2} provides another case study using the REDDIT-BINARY social network dataset in three different configuration scenarios, where our \gvex explanation view successfully determines representative patterns for different labels of interest. The three configuration scenarios indicate whether the user prefers only one class or is interested in the nature of both classes. For {\em online-discussion} threads, user interactions typically resemble star-like structures, where many strangers post their thoughts on a popular topic. Our explanation pattern $P_{61}$ aids in distinguishing these topic groups within explanation subgraphs. 
On the other hand, in {\em question-answer} threads, users exhibit biclique-like patterns $P_{81}$, capturing the phenomenon where a few domain experts actively provide answers to various questions raised by different users in closely related domains. When user attempts to understand both classes, \gvex presents the salient patterns of both classes, shedding light on  important patterns that underpin the classification of this social network dataset. 
\vspace{-2mm}
\section{Conclusion}
\label{sec:concl}
We proposed \gvex, a novel 
graph view-based two-tier structure to 
explain \gnn-based graph classification. 
We established hardness results 
for explanation view generation, and provided efficient algorithms with provable 
performance guarantees. We 
experimentally verified 
that \gvex-based explaination 
outperforms existing 
techniques in terms of 
conciseness, explanability, and 
efficiency. Our algorithms show good performance on different domains: (social networks, chemistry, biology) and types: (directed/undirected, sparser/denser, with/without node features) of graphs, considering both binary and multi-class classification problems, under static and streaming settings.
In future, we shall consider the impact of edge features and develop distributed view-based \gnn explanation.

\begin{acks}
Tingyang Chen, Xiangyu Ke, and Yunjun Gao are supported in part by the NSFC under Grants No. (62025206, U23A20296) and Yongjiang Talent Introduction Programme (2022A-237-G). Dazhuo Qiu and Arijit Khan acknowledge support from the Novo Nordisk Foundation grant NNF22OC0072415. Yinghui Wu is supported in part by NSF under CNS-1932574, ECCS-1933279, CNS-2028748 and OAC-2104007. Xiangyu Ke is the corresponding author.
\end{acks}

\bibliographystyle{ACM-Reference-Format}
\bibliography{ref}


\appendix

\section{Appendix: Proofs, Algorithms \& 
Experimental Study}

\subsection{Proof of Lemma~\ref{lm-verifynpc}}
{\em  Given a graph database $\G$, configuration $\C$, and a two-tier structure $(\P, \G_s)$, the view verification problem is \NP-complete when the \gnn $\M$ is fixed.}

\begin{proof}
It is not hard to verify that view verification is \NP-hard, given that 
it requires subgraph isomorphism 
tests alone to verify constraint 
\textbf{C1}, which is known 
to be \NP-hard~\cite{floderus2015induced}.  

We next outline an \NP algorithm for the verification problem. 
It 
performs a three-step 
verification below. 
(1) For \textbf{C1}, it guesses a finite number of matching  
functions in \PTIME 
(for patterns $\P$ and $\G$ with bounded size), and verifies if the patterns induce 
accordingly $\G_s$ via the matching 
functions in \PTIME. If so, 
$\G_\V$ is a graph view. 
(2) To check \textbf{C2}, for 
each graph $G\in \G$ and 
its corresponding 
subgraphs $G_s\in\G_s$, 
it applies $\M$ to verify 
if $\M(G_s)$ = $l$ and 
$\M(G\setminus G_s)\neq l$. 
If so, $\G_\V$ is an 
explanation view for $\G$. 
For a fixed \gnn $\M$, 
it takes \PTIME to do the 
inference. 
(3) It takes \PTIME to 
verify the coverage given 
that subgraph isomorphism 
tests have been performed 
in steps (1) and (2). 
These verify the upper bound 
of the verification problem. 
\end{proof}

\subsection{Proof of Theorem~\ref{thm-hardness}} 
{\em For a fixed \gnn $\M$, 
\evg is (1) $\Sigma^2_P$-complete, and 
(2) remains $NP$-hard even 
when $\G$ has no edges.  } 

\begin{proof}
    We first show that \evg is solvable in $\Sigma^2_P$. We set an \NP oracle for view verification, which calls the \NP algorithm in 
the proof of Lemma~\ref{lm-verifynpc} to check if a pair $(\P, \G_s)$ satisfies the three 
constraints to be an explanation view under 
the configuration $\C$ for a single 
label $l\in \L$ and a single 
graph $G\in \G$. We outline a second \NP algorithm below that consults the 
above \NP oracle. 
The nondeterministic algorithm 
guesses a set of two-tier view structures 
\gvl = $\{(\P,\G_s)_i\}$ ($i\in[1,|\L|]$), and 
determines if for each label group 
$\G^l$, it contains 
an explanation view $(\P^l,\G^l_s)$, 
by calling the above \NP oracle, in 
$O(|\L||\P||\G|)$ time. If so, 
it then computes $f(\G^l_\V)$ 
and checks if 
$f(\G^l_\V)\geq h$ in 
\PTIME. 

\sstab
(2) To see that \evg is $\Sigma^2_P$-hard, 
we construct a reduction from graph satisfiability, a known $\Sigma^2_P$-complete problem~\cite{schaefer2002completeness}. 
Given two sets $\G^+$ and $\G^-$ 
of graphs with labels '+' and '-' respectively, graph satisfiability problem  
determines whether there exists a graph $G_o$ 
such that each graph $G^+\in \G^+$ is 
isomorphic to a subgraph of $G_o$, 
and each $G^-\in \G^-$ is not isomorphic 
to any subgraph of $G_o$. Our 
reduction assumes that a fixed \gnn $\M$ as a 
binary classifier is provided, 
and performs a preprocessing step in 
\PTIME as follows. 
(i) Given an instance of 
graph satisfiability, we first apply 
$\M$ to $\G^+$ and $\G^-$ and 
``regroup'' them into two new 
groups $\G^+_\M$ and $\G^-_\M$, according to the result of $\M$. 
(ii) We then
augment $\G^+_\M$ (resp. $\G^-_\M$) 
into a new set $\G_\M^{+'}$ (resp. $\G_\M^{-'}$), where for
each graph $G_i^+\in\G_\M^+$ (resp. $G_j^-\in\G_\M^-$), a single independent  
node $v^-_i$ (resp. $v^+_j$) with a class
label '-' (resp. '+') verified by $\M$ 
is added, \ie $\M(v_i)$ = '-'
(resp. $\M(v_j)$ = '+'). Such nodes 
can be obtained with $\M$ inference
over all the single nodes (as independent nodes) in $\G$, hence in \PTIME. 
We set graph database 
$\G$ = $\G_\M^{+'}\cup\G_\M^{-'}$. 
(3) We set in $\C$ the coverage constraints 
$[|\G_\M^{+'}|, |\G_\M^{+'}|]$ for label '+' (resp. $[0,0]$ for '-'). One can verify that 
there exists a 
solution for graph satisfiability if and only if there is an explanation view for $\G$ 
that satisfies $\C$. 

\sstab 
(3) To see Theorem~\ref{thm-hardness}(2), we consider 
a special case of \evg. Let $\G$ contains two single graphs $G_1$ and $G_2$, 
each has no edge. A pre-trained \gnn $\M$ 
as a binary classifier assigns labels on graph nodes (\ie $\L$ contains two labels). For such a case, 
\evg remains to be \NP-hard. To see this, we construct a reduction 
from the red-blue set cover problem~\cite{robert00}, which is \NP-complete. 
This verifies the hardness of \evg for identifying
explanation with coverage requirement alone, as in such case, subgraph isomorphism test is no longer intractable.
\end{proof}
\vspace{-1.5ex}
\subsection{Proof of Lemma~\ref{lm-submodular}}
{\em Given $\G$, $\L$, $\C$ and a fixed \gnn $\M$,  $f(\G_\V)$ is a monotone submodular function. }

\begin{proof}
As $f(\G_\V)$ is the sum of 
$f(\G^l_\V)$, where $l$ ranges over $\L$, 
and (1) each $f(\G^{l}_{\V})$ 
is the sum of a node set function $f'(V_{si})$ for each graph 
$G_i$ in label group $\G^l$, 
and (2) each $f'(V_{si})$ is in turn 
only determined by two component node set functions 
 $I(V_{si})$ and $D(V_{si})$, 
 one only needs to show that both
its components $I(V_{s})$ and $D(V_{s})$ are monotone submodular (see Equation \ref{eq:goodness}). 

A function $f'(V_s)$ is
submodular if for any subsets $V_{s''}\subseteq V_{s'} \subseteq V_s$ and any node
$u \notin V_{s'}$, (i) $f'(V_{s''}) \leq f'(V_{s'})$,  and (ii) $f'(V_{s''} \cup \{u\}) - f'(V_{s''}) \ge f'(V_{s'}\cup \{u\}) - f'(V_{s'})$~\cite{calinescu2011maximizing}.
   
\sstab
(1) We first show that $I(\cdot)$ is monotone submodular. Given the node set $V_s$, 
we denote as $\kw{Inf}(V_s)$ 
the node set influenced by $V_s$ 
\wrt thresholds $(\theta, r)$ 
(as specified in configuration $\C$); \ie 
$I(V_s)$ = $|\kw{Inf}(V_s)|$. (a) Clearly, 
for any subset $V_{s''}\subseteq V_{s'}$, 
$\kw{Inf}(V_{s''})\subseteq\kw{Inf}(V_s')$, thus $I(V_{s'})$= 
$|\kw{Inf}(V_{s'})| \ge |\kw{Inf}(V_{s''})|$=$I(V_{s'})$. 

(b) To see its submodularity, 
we next show that for any set 
$V_{s''}\subseteq V_{s'}$ and 
any node $u\not\in V_{s'}$, 
\begin{equation}
    |\kw{Inf}(V_{s''} \cup \{u\})| - |\kw{Inf}(V_{s''})| \ge |\kw{Inf}(V_{s'} \cup \{u\})| - |\kw{Inf}(V_{s'})|
\end{equation}

It then suffices to show that 
$|\kw{Inf}(V_{s'}\cup\{u\})|$ - 
$|\kw{Inf(V_{s''}\cup\{u\})}|\leq$ 
$|\kw{Inf}(V_{s'})|$-$|\kw{Inf}(V_{s''})|$.
Note that $u\not\in V_{s'}$, and $u\not\in V_{s''}$. 
Thus $|\kw{Inf}(V_{s'}\cup\{u\})|$ - 
$|\kw{Inf}(V_{s''}\cup\{u\})|$ 
= $|\kw{Inf}(V_{s'})\cup \kw{Inf}(\{u\})|$
- $|\kw{Inf}(V_{s''})\cup \kw{Inf}(\{u\})|$. 
(i) If $\kw{Inf}(\{u\})\cap \kw{Inf}(V_{s'})$ 
= $\emptyset$, then 
we have the above equation trivially 
equals $|\kw{Inf}(V_{s'})|$ + $|\kw{Inf}(\{u\})|$
- $(|\kw{Inf}(V_{s''})$ + $|\kw{Inf}(\{u\})|$)
= $|\kw{Inf}(V_{s'})|$ - $|\kw{Inf}(V_{s''})|$. 
(ii) Otherwise, $\kw{Inf}(\{u\})\cap \kw{Inf}(V_{s'})\neq\emptyset$. 
Note that $|\kw{Inf}(V_{s'})|$-
$|\kw{Inf}(V_{s''})|$ = 
$|\kw{Inf}(V_{s'})\setminus\kw{Inf}(V_{s''})|$. Then, 
$|\kw{Inf}(V_{s'}) \cup \kw{Inf}(\{u\})|$ - $|\kw{Inf}(V_{s''}) \cup  \kw{Inf}(\{u\})|$ = $|(\kw{Inf}(V_{s'})\setminus\kw{Inf}(V_{s''})) \setminus \kw{Inf}(\{u\})| 
\le |\kw{Inf}(V_{s'})\setminus\kw{Inf}(V_{s''})|$. 
Putting these together, 
the submodularity of $I(\cdot)$ hence follows. 

\sstab 
(2) Following a similar analysis, we can show that $D(V_s)$ is also monotone submodular. 
As both $I(V_s)$ and $D(V_s)$ are monotone submodular, and the sum of monotone 
submodular functions remain to be 
monotone submodular, Lemma~\ref{lm-submodular} follows.    
\end{proof}
\vspace{-1.5ex}
\subsection{Proof of Lemma \ref{lm-coverage}}

{\em For a given set of explanation subgraphs $\G^l_s$, 
procedure~\psum is an $H_{u_l}$-approximation of 
optimal $\P^l$ that ensures 
node coverage (hence satisfies coverage constraint 
in $\C$). Here, $H_{u_l}$ = $\sum_{i\in[1,\C.u_l]}\frac{1}{i}$ is the $u_l$-th Harmonic number ($\C.u_l\geq 1$).}

\begin{proof} 
We show the optimality guarantee by performing a 
reduction to the minimum weighted set cover problem (MWSC). 
The problem of MWSC takes as input a universal set 
$X$ and a set of weighted subsets $\X$= $\{X_1, \ldots, X_n\}$. 
Each subset $X_i\in \X$ has a weight $w_i$. The problem 
is to select up to $k$ subsets $\X_k\subseteq\X$
= $\{X_1, \ldots X_k\}$ such that $X$ = $\bigcup_{j\in[1,k]} X_j$, with a minimized total sum of weights. 
(1) Given a set of explanation subgraphs $\G^l_s$, 
we set the union of the nodes $V_\S$ as $X$. 
we consider the pattern candidates $\P$ generated from 
procedure \pgen. For each pattern $P_i\in\P$, 
we set the node set $P_{V_S}$ that are covered by $P$ in 
$V_\S$ as a subset $X_i$, and associate the 
number of uncovered edges in $\G^l_s$ as $w(P_i)$. 
This transforms our problem to an instance of 
an MWSC problem. 
(2) Given a solution $X_k$, we 
simply set $\P^l$ to the set of patterns that 
are corresponding to the selected subsets in $X_k$.  
This transforms the solution back to 
the solution to our problem. 
Then we can readily verify the following. 
(a) The above constructions are in \PTIME (in terms of 
input sizes). 
(2) Assume there exists a solution 
$X_k$ that approximates an optimal solution 
$X*_k$ for MWSC with ratio $\alpha$, 
then the corresponding 
solution $\P^l$ is an $\alpha$-approximation 
for our problem. This is because 
the weights are consistently defined 
as the edge cover loss for each pattern 
independently. Given the above analysis, 
Lemma~\ref{lm-coverage} follows. 
\end{proof}
\vspace{-1.5ex}
\subsection{Procedure IncUpdateVS}

\kw{IncUpdateVS} consults a greedy swapping strategy to decide whether to replace a node $v'\in V_S$ with $v$ or reject $v$ and put the node $v'$ into $V_u$. It performs a case analysis: (a) if it can simply adds $v$ to $V_S$; (b) otherwise, if $\P^l$ already 
covers $v$, or $v$ alone does not contribute new patterns to $\P^l$ ($\Delta\P = \emptyset$, as determined by invoke \kw{IncPGen}), it skips processing $v$; (c) otherwise, it chooses the node $v'\in V_S$ whose removal has the smallest ``loss'' of explainability score (Line 1) and replaces $v'$ with $v$ only when such a replacement ensures a gain that is at least twice as much as the loss (Line 2-5). The detail of case(c) is shown in Procedure \ref{procedure:incupdatevs}. 
\floatname{algorithm}{Procedure}
\begin{algorithm}[htb!]
    \renewcommand{\algorithmicrequire}{\textbf{Input:}}
    \renewcommand{\algorithmicensure}{\textbf{Output:}}
    \caption{\kw{IncUpdateVS}$(v,V_S,V,G,G_s)$}
    \begin{algorithmic}[1]
        \STATE $v^- := \mathop{\mathrm{argmin}}_{v' \in V_S} (f(V_S)-f(V_S \backslash v'))$
        \STATE $V_u := V_S \backslash \{v^-\}$
        \STATE $w(v) = f(V_u \cup v)-f(V_u)$;$w(v^-) = f(V_u \cup v^-)-f(V_u)$
        \IF{$w(v) \ge 2w(v^-)$}
        \STATE $V_S := V_S \backslash \{v^-\} \cup \{v\}$
        \ENDIF
        \RETURN $V_S$
    \end{algorithmic}
    \label{procedure:incupdatevs}
\end{algorithm} 
\vspace{-1.5ex}
\subsection{Procedure IncUpdateP}

\kw{IncUpdateP} performs a similar case analysis, yet on patterns $\P^l$, and conducts a swapping strategy to ensure node coverage and 
small edge misses. For newly maintained $V_S$, first, ensure meeting node coverage constraints (Line 4-8) by generating new patterns based on the unseen induced explanation subgraph (Line 9-11); second, based on the normalized weight $w(P)$ (Line 12), swapping patterns that have no contribution to the node coverage and have the biggest edge misses (Line 13-14). The detail is shown in Procedure \ref{procedure:incupdatep}. 

\floatname{algorithm}{Procedure}
\begin{algorithm}[htb!]
    \renewcommand{\algorithmicrequire}{\textbf{Input:}}
    \renewcommand{\algorithmicensure}{\textbf{Output:}}
    \caption{\kw{IncUpdateP}$(v,V_S,\P_c)$}
    \begin{algorithmic}[1]
        \STATE set $\P^\prime := \emptyset$; $v$ induced subgraph $G_{s_v}$; and corresponding node set $V_v$;
        \FOR{$v^\prime \in V_S$}
        \STATE $U := \emptyset$
        \FOR{$P \in \P_c$}
        \IF{$P$ and $G_{s_{v^\prime}}$ are isomorphic}
        \STATE $U := U \cup \{V_P\}$
        \IF{$P \notin \P^\prime$}
        \STATE $\P^\prime := \P^\prime \cup \{P\}$
        \ENDIF
        \ENDIF
        \ENDFOR
        \IF{$U \neq V_{v^\prime}$}
        \STATE $P_{new} := G_{s_{v^\prime}}(V_{v^\prime}\backslash U)$
        \STATE $\P^\prime := \P^\prime \cup P_{new}$
        \ENDIF
        \ENDFOR
        \STATE $w(P) = 1-|P_{E_S}| \backslash |E_S|$
        \STATE $\P^- \in \P_c\backslash \P^\prime$  and has biggest $w(\P^-)$
        \STATE $\P_c := \P_c\backslash \P^- \cup \P^\prime$
        \RETURN $\P_c$
    \end{algorithmic}
    \label{procedure:incupdatep}
\end{algorithm} 

\vspace{-1.5ex}
\subsection{Parallel Implementation}
Within our algorithm, the calculation of {\em Feature Influence} and {\em Neighborhood Diversity} for each graph is carried out independently. This observation presents a valuable opportunity for the parallelization of our algorithm. Consequently, we are not constrained to relying on a single process to handle all the graphs simultaneously. By employing multi-process execution on a $48$-core CPU, we can efficiently distribute the computational load among multiple processes, allowing each process to compute the respective graph autonomously. This approach enables us to enhance the efficiency of the \gvex algorithm. Additionally, these similar concepts can be readily extended to distributed systems.

\begin{figure}[ht!]
	\vspace{-2.5mm}
	\centering
	\subfigure[Explanation views for different node orders]
	{\includegraphics[width=0.52\linewidth]{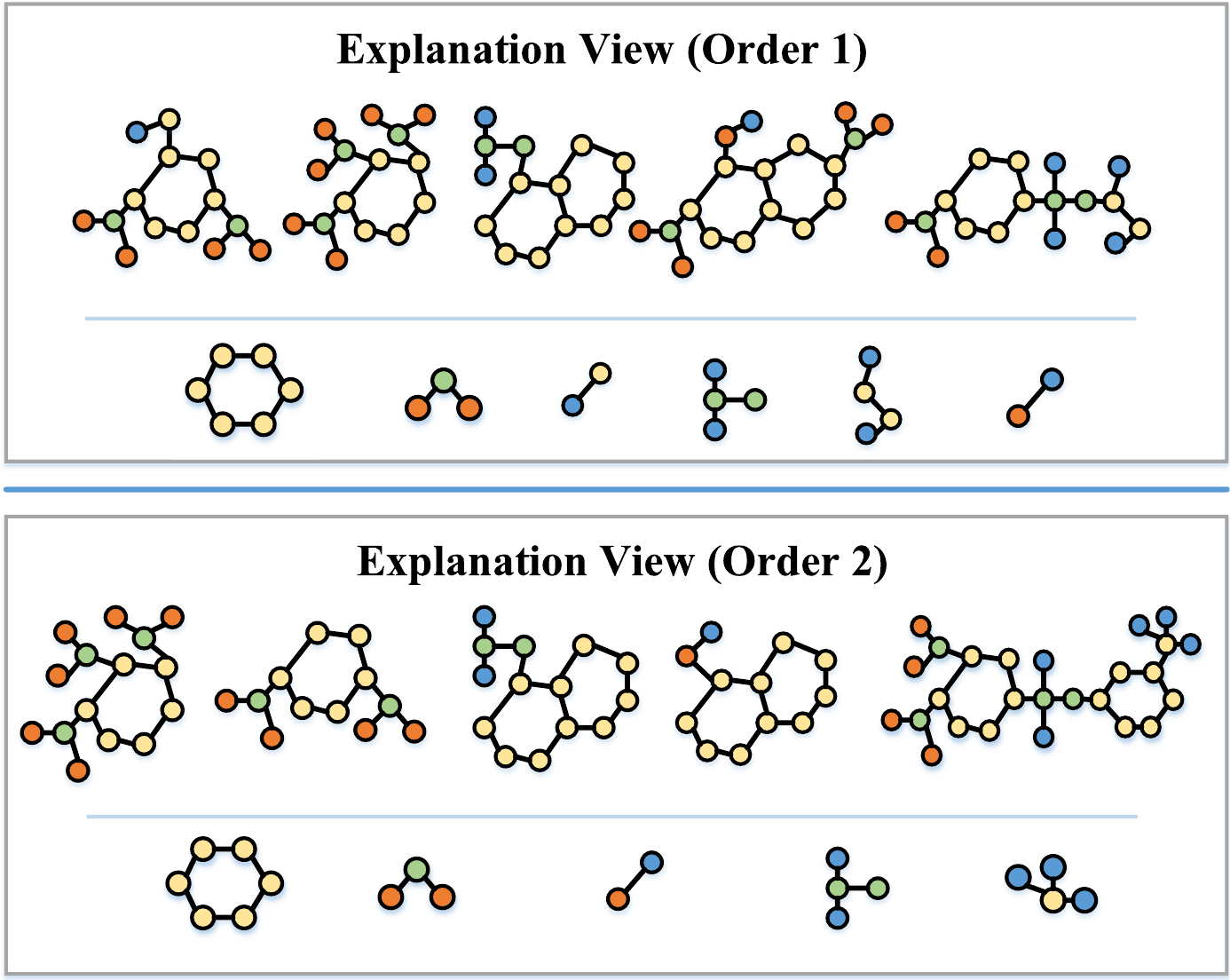}
		  \label{fig:random-sg-views}
         }
	\subfigure[Runtime for different node orders]
	{\includegraphics[width=0.45\linewidth]{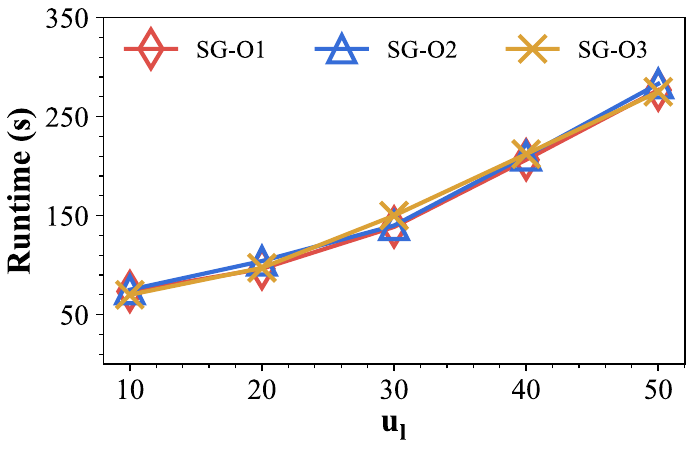}
		  \label{fig:random-sg-time}
         }
	\vspace{-3mm}
	\caption{Different node orders in \gvstream, {\sf MUT} dataset}
        \label{fig:random-sg}
	\vspace{-3mm}
\end{figure}

\subsection{Node order analysis \wrt \gvstream.}
The streaming setting does not require a predefined order of nodes. \gvstream ascertains an "anytime" quality guarantee, regardless of the node sequences (Theorem 5.1). Our approximation ratio holds w.r.t. an optimal explanation view on the ``seen''  fraction of $\G$, thus, providing a pragmatic solution for large $\G$. The node arrival sequence inherently impacts the order of pattern discovery, potentially resulting in the early identification of certain patterns. Furthermore, due to our replacement strategies such as {\sf IncUpdateVS} and {\sf IncUpdateP}, the arrangement of higher-tier patterns may undergo slight modifications. These strategies intelligently oversee the management of patterns within $\mathcal{P}^l$ through efficient "swapping", allowing for real-time decisions on patterns and node replacement. Consequently, subtle variations may arise in higher-tier patterns contingent upon distinct node processing orders. However, given the approximation guarantee within our algorithm, coupled with the continuous update of pattern information, the vast majority of crucial patterns will persist, even under varying node orders, thus exhibiting minimal alterations in the ultimate result.
Our additional example (Figure~\ref{fig:random-sg-views})  illustrates that there exists a slight difference in the higher-tier patterns from those shown in Figure~\ref{fig-alg-stream} under different processing orders. Notably, node orders do not affect the worst-case time cost of \gvstream. Figure \ref{fig:random-sg-time} validates similar running times on the {\sf MUT} dataset for various node execution orders obtained via random shuffles.

\begin{figure}[ht!]
	\vspace{2mm}
	\centering
	\includegraphics[width=0.8 \linewidth]{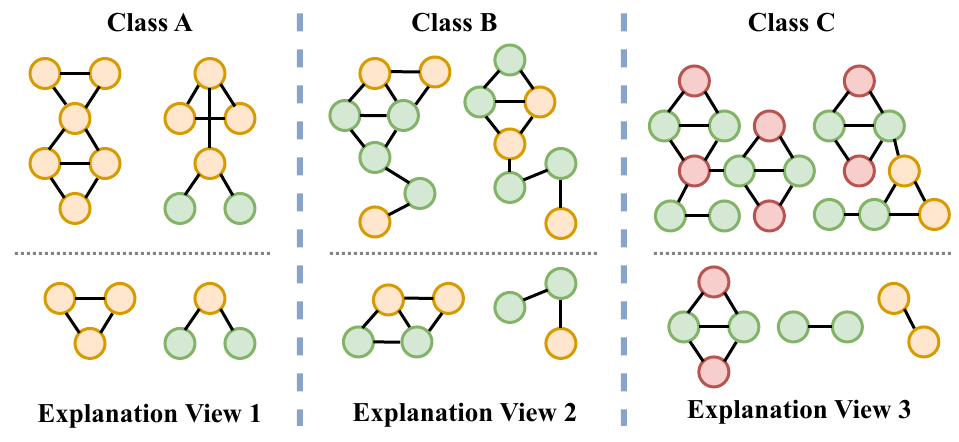}
        \vspace{-3mm}	
	\caption{Explananion views on {\sf ENZ} dataset}
	\label{fig:case-enzymes}
   \vspace{-5mm}
\end{figure}

\subsection{Case study on ENZYMES.}
 We further extend the current case studies. We added an analysis of the {\sf ENZ} dataset (biology), from which three classes are taken out as examples for the generation of the explanation views (Figure ~\ref{fig:case-enzymes}).  This shows that the proposed methods are effective in terms of identifying different subgraph structures.

\end{document}